\documentclass{article}
\usepackage[english]{babel}
\usepackage[letterpaper,top=2cm,bottom=2cm,left=3cm,right=3cm,marginparwidth=1.75cm]{geometry}
\usepackage{amsmath}
\usepackage{graphicx}
\usepackage[colorlinks=true, allcolors=blue]{hyperref}
\usepackage{booktabs}
\usepackage{amsmath}
\usepackage{xcolor}
\usepackage{listings}
\usepackage{subcaption}
\usepackage{pgfplots}
\pgfplotsset{compat=1.18}
\usepackage{tikz}

\usepackage[final,nonatbib]{neurips_2023}
\usepackage[utf8]{inputenc}
\usepackage{multicol}
\newcommand{\osa}{{ODS}} 
\newcommand{\osasp}{{ODS }}

\title{Open Deep Search: Democratizing Search with Open-source Reasoning Agents}

\makeatletter
\renewcommand*{\@fnsymbol}[1]{\ensuremath{\ifcase#1\or \dagger\or \dagger\or \ddagger\or
   \mathsection\or \mathparagraph\or \|\or **\or \dagger\dagger
   \or \ddagger\ddagger \else\@ctrerr\fi}}
\makeatother

\author{
  Salaheddin Alzubi$^\dagger$ \\
 \And
  Creston Brooks$^\dagger$ \\
  \And
  Purva Chiniya$^\dagger$ \\
  \And
  Edoardo Contente$^\dagger$
  \And
  Chiara von Gerlach$^\dagger$   \And 
  Lucas Irwin$^\dagger$  \And
  Yihan Jiang$^\dagger$ 
  \And 
  Arda Kaz$^{\dagger\star}$  
 \And
 Windsor Nguyen$^{\dagger\ddagger}$
 \And
  Sewoong Oh$^{\dagger *}$ \\
  \And Himanshu Tyagi$^\dagger$ \And Pramod Viswanath$^\ddagger$\thanks{Sentient,\;\; $^*$University of Washington,\;\;$^\ddagger$Princeton University,\;\;$^\star$UC Berkeley}  
}
\date{} 

\begin{document}
\maketitle

\begin{abstract} 
We introduce Open Deep Search (ODS) to close the increasing gap between the proprietary search AI solutions, such as Perplexity's Sonar Reasoning Pro and OpenAI's GPT-4o Search Preview, and their open-source counterparts. The main innovation introduced in ODS is to augment the reasoning capabilities of the latest open-source LLMs with reasoning agents that can judiciously use web search tools to answer queries. Concretely, ODS consists of two components that work with a base LLM chosen by the user: Open Search Tool and Open Reasoning Agent. Open Reasoning Agent interprets the given task and completes it by orchestrating a sequence of actions that includes calling tools, one of which is the Open Search Tool. Open Search Tool is a novel web search tool that outperforms proprietary counterparts. Together with powerful open-source reasoning LLMs, such as DeepSeek-R1, ODS nearly matches and sometimes surpasses the existing state-of-the-art baselines on two benchmarks: SimpleQA and FRAMES. For example, on the FRAMES evaluation benchmark, ODS improves the best existing baseline of the recently released GPT-4o Search Preview by 9.7\% in accuracy. ODS is a general framework for seamlessly augmenting any LLMs---for example, DeepSeek-R1 that achieves 82.4\% on SimpleQA and 30.1\% on FRAMES---with search and reasoning capabilities to achieve state-of-the-art performance: 88.3\% on SimpleQA and  75.3\% on FRAMES.

\end{abstract}

\section{Introduction}

Search AIs--Search engine-augmented Large Language Models (LLMs)--combine the Retrieval Augmented Generation (RAG) capabilities of LLMs (e.g., \cite{lewis2020retrieval}) with real-time information retrieval from search engines. This integration addresses the challenge of LLMs' static knowledge base enabling LLMs to provide up-to-date and contextually relevant responses. 
Recent works like \cite{vu2023freshllmsrefreshinglargelanguage}
have shown that feeding Search Engine Result Page (SERP) APIs as context to an LLM outperforms previous methodologies, such as self-ask.

Advances in search AI have been dominated by proprietary solutions such as Google search, Bing, chatGPT search, and Grok. In particular, Perplexity AI \cite{perplexity2024} has excelled in this market, even threatening mature leaders. However, such closed-source solutions limit transparency, innovation, and entrepreneurship. With the goal of growing a community of developers in search AI, harnessing collective talent, fostering innovation, and encouraging entrepreneurship, we introduce Open Deep Search (\osa), an open-source AI search solution that achieves state-of-the-art performance in benchmark evaluations, matching or surpassing those achieved by the best closed-source alternatives.

\begin{table}[h!]
\centering
\renewcommand{\arraystretch}{1.2}
\begin{tabular}{c|l|cc}
\hline
 & & \textbf{SimpleQA} & 
 \textbf{FRAMES} \\
\hline
&Llama3.1-70B & 20.4 
& 34.3 \\
w/o access to the web & DeepSeek-R1 & 82.4 
& 30.1 \\ 
& GPT-4o & 37.5 
& 50.5 \\
\hline
& Perplexity~\cite{perplexity2024} & 82.4 
& 42.4 \\
& Perplexity Sonar Reasoning Pro \cite{perplexity_sonar_pro}& 85.8 
& 44.4 \\
with access to the web & GPT-4o Search Preview \cite{gptsearch} & \textbf{90.0} & 65.6 \\
& \osa-v1+DeepSeek-R1 (ours) & 87.7 & 56.7 \\
& \osa-v2+DeepSeek-R1 (ours) & 88.3 & \textbf{75.3}  \\
\hline
\end{tabular}
\vspace{0.5cm} 
\caption{The proposed open-source search framework of \osa, when used with the open-source reasoning LLM of DeepSeek-R1 \cite{deepseek}, achieves performance exceeding that of closed-source state-of-the-art search AI solutions of Perplexity \cite{perplexity2024}, Perplexity Sonar Reasoning Pro \cite{perplexity_sonar_pro} on the two popular factuality evaluation Benchmarks of FRAMES \cite{krishna2024} and SimpleQA \cite{wei2024measuring}. Compared to GPT-4o Search Preview, \osa-v2 has a significantly better FRAMES accuracy but a slightly worse SimpleQA accuracy. \osa-v1 uses a ReAct-based agent (Section~\ref{sec:ODS-v1}) and \osa-v2 uses a CodeAct-based agent (Section~\ref{sec:ODS-v2})}.
\label{tab:main}
\end{table}

Perplexity AI has two search AI products with API accesses: the default Perplexity \cite{perplexity2024} and Perplexity Sonar Reasoning Pro tailored for complex reasoning tasks \cite{perplexity_sonar_pro}. Table~\ref{tab:main} shows their performance on two popular evaluation benchmarks of SimpleQA \cite{wei2024measuring} and FRAMES \cite{krishna2024}. Note that both versions of the proposed Open Deep Search (when used together with the open-source DeepSeek-R1 model) outperform the Perplexity AI's flagship search AI products. Another important baseline is OpenAI's GPT-4o Search Preview \cite{gptsearch} tailored for search AI and released on 3/11/2025. The second version of Open Deep Search, dubbed \osa-v2+DeepSeek-R1, surpasses GPT-4o Search Preview on FRAMES and almost matches it on SimpleQA. To achieve this state-of-the-art performance, we make two innovations in Open Deep Search:  Open Search Tool and Open Reasoning Agent.



\medskip\noindent{\bf Open Search Tool.}
Perplexity's  and OpenAI's search solutions are closed-source products. Open-source alternatives to Perplexity, such as OpenPerplex \cite{openperplex} and Perplexica \cite{perplexica} are open-source search tools whose output is summarized and fed into an LLM to answer the query of interest.  However, these open-source search tools have several weaknesses that leave significant room for improvement. First, OpenPerplex and Perplexica primarily pass raw SERP results as context to the LLM. Additionally, they chunk and rerank snippets based on their relevance to the user query. In contrast, our approach employs a more sophisticated search process, which we call Open Search Tool (explained in detail in Section~\ref{sec:OpenSearchTool}). Specifically, we rephrase queries when necessary, extract context from the top $N$ snippets, and apply chunking and re-ranking to filter content above a relevance threshold. This ensures the inclusion of all relevant search result context. Additionally, we implement custom website handling for major APIs such as Wikipedia, ArXiv, and PubMed.

\medskip\noindent{\bf Open Reasoning Agent.} As shown in Table~\ref{tab:main}, we provide two versions of Open Reasoning Agent: one based on the ReAct agent \cite{yao2022react} and another based on the CodeAct agent \cite{lv2024codeact}. 
The goals of Open Reasoning Agent are to interpret the query, assess the retrieved context, and use appropriate tools, including Open Search Tool we provide for web search, to answer the query. For details we refer to Section~\ref{sec:OpenReasoningAgent}.  

\medskip\noindent{\bf Outline.} In the next section, we give details of the search AI that we call Open Deep Search. In Section~\ref{sec:results}, we provide empirical results on the two evaluation benchmarks of SimpleQA and FRAMES, along with an ablation study of various components of ODS.

\section{Open Deep Search (\osa)} 

Open Deep Search (\osa) is a plug-and-play framework, where the user can seamlessly plug in any base LLM of their choice, either open-source LLMs or through API accesses to closed-source LLMs. In our experiments, we use \osasp with the Llama3.1-70B model or with DeepSeek-R1. 
\osasp consists of two parts, Open Search Tool and Open Reasoning Agent, both fo which use the base LLM. Open Search Tool is our open-source tool for searching and processing the information from the Internet explained in Section~\ref{sec:OpenSearchTool} and Open Reasoning Agent is our open-source agent that answers the queries with access to tools as explained in Section~\ref{sec:OpenReasoningAgent}. 

We provide two solutions for the Open Reasoning Agent: one using the ReAct agent (which we call \osa-v1 for version one) and one using the CodeAct agent (which we call \osa-v2 for version 2).
The open-source repository that contains both versions of ODS can be found at \url{https://github.com/sentient-agi/OpenDeepSearch}.
We use the terminology \osa-v?+base-model to refer to a specific instantiation for \osa. For example, \osa-v1+Llama3.1-70B refers to the version one of \osasp that uses the ReAct agent and uses Llama3.1-70B as the base model, and \osa-v2+DeepSeek-R1 refers to the version one of \osasp that uses CodeAct agent and DeepSeek-R1 as the base model.



\begin{figure}[h!]
    \centering
    \includegraphics[width=.7\linewidth]{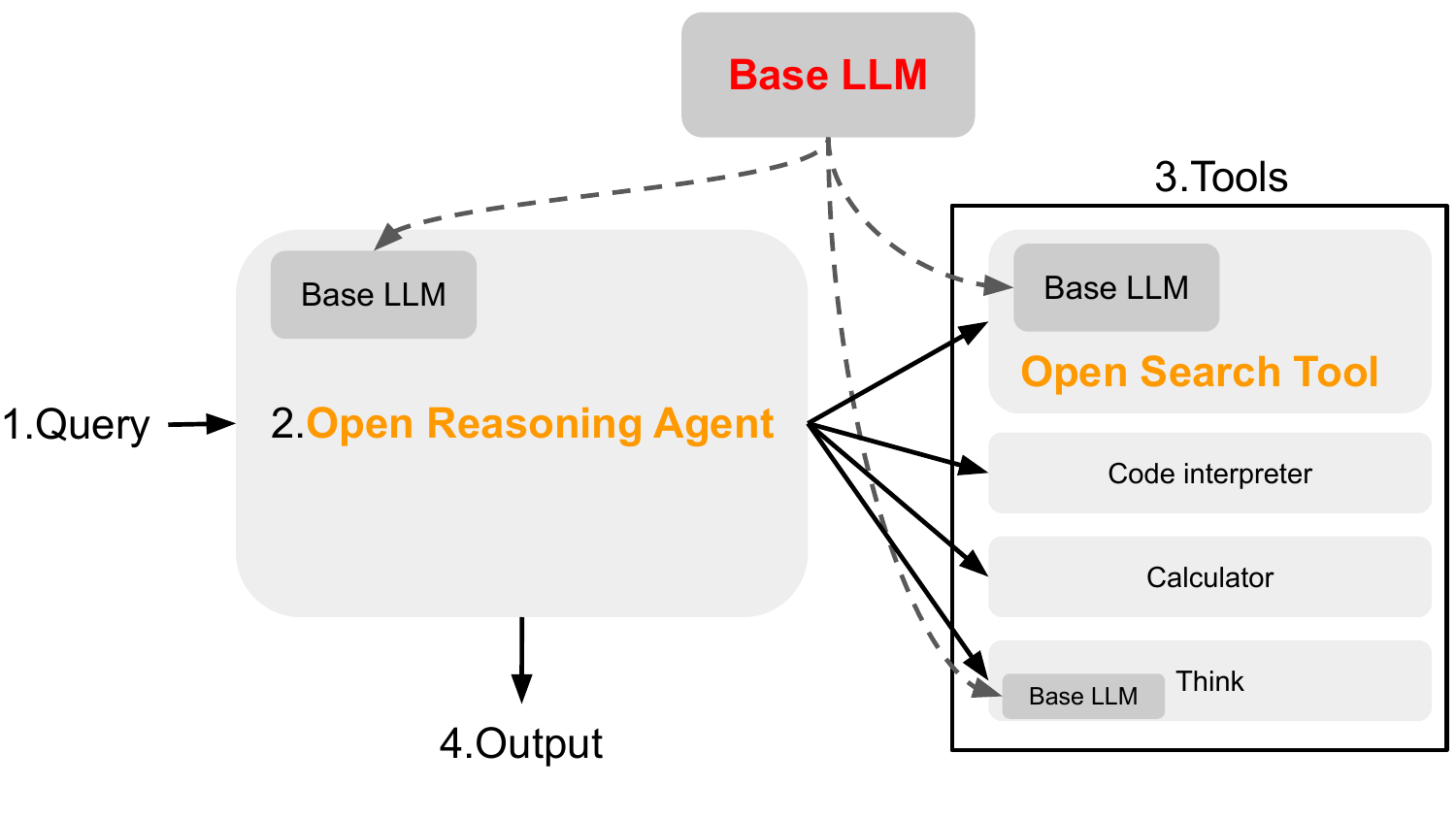}
    \caption{A user can choose to plug in any base LLM of their choice and harness the benefits of the open-source framework of Open Deep Search (\osa),  which consists of two components: Open Search Tool and Open Reasoning Agent. A query is first fed into Open Reasoning Agent which orchestrates the available set of tools to interpret and answer the query. The most important tool is the Open Search Tool that we design, which provides a high quality context from multiple retrieved sources from the web. In our experiments we use Llama3.1-70B and DeepSeek-R1 as our base model.}
    \label{fig:system_architecture}
\end{figure}

\subsection{Open Search Tool} 
\label{sec:OpenSearchTool} 

We provide Open Search Tool that improves upon recent advancements \cite{vu2023freshllmsrefreshinglargelanguage} in search engine augmented LLM's whilst maintaining the commitment to a \textbf{scalable open-source} system. In particular, our approach revisits different components along the querying, retrieval, and augmentation pipeline of conventional methodologies. Open Search Tool takes a  query as input and produces a context to be fed into the base LLM that consists of relevant passages from the web search. The quality of the Open Search Tool is critical in the success of \osa, as we show with examples in Section~\ref{sec:results} and Appendix~\ref{app:examples}. 

\subsubsection{Query Rephrasing}

The first step in our search pipeline involves ingesting the original user query and generating $k$ new rephrased queries that maintain the original context. The necessity of this step stems from the gap between the original query's semantic structure and the underlying context sought to provide a satisfactory answer. For example, a user may be interested in knowing ``how to make my Internet faster", however, a Google search may not yield satisfactory results as the query itself is too broad while the implicit context may not be covered. The query rephraser bridges that gap between broad queries and implicit context and will generate other queries such as: ``How to make the Wi-Fi Signal stronger", ``How to increase bandwidth", and ``How to reduce latency". We have found this step to be crucial in improving the coverage and diversity of retrieved contexts, and consequently, the overall performance of our system. 

\subsubsection{Retrieval}

The second step in our search pipeline involves retrieving relevant contexts from a search engine result page API (SERP)\footnote{In our case, we used: serper.dev}. The retrieved results from the API call are then formatted, processed, and inserted into the context of an LLM. Our formatting procedure is inspired by the \textbf{FreshPrompt} \cite{vu2023freshllmsrefreshinglargelanguage} prompt format where we include meta-data returned each search result such as the title, URL, the description, and date authored (if available) from every snippet. Additionally, we prompt the LLM to prioritize reliable sources, such as government agencies, educational institutions, and established research institutions, over other sources when the search context contains conflicting information. 

\subsubsection{Augmentation}

As an additional step in our pipeline, we choose to augment the LLM's context by adding relevant passages from the top $m$ retrieved links by the SERP API. In particular, we scrape the associated webpages, embed passage chunks and retrieve the top $n$ relevant passages from each webpage based on the re-ranker score assigned relevant to the user query. This enables the contexts to provide in-depth answers related to queries that elicit ``deeper" reasoning.

\subsection{Open Reasoning Agent}
\label{sec:OpenReasoningAgent}


Open Reasoning Agent takes a query from the user as an input and uses the base LLM and various tools to produce an answer. 
The Open Search Tool from the previous section is one of the critical tools for Open Reasoning Agent to use. We provide two solutions for the user to choose from: one that is based on Chain-of-thought and ReAct agent (whose resulting end-to-end system is referred to as \osa-v1) and one that is based on Chain-of-code and CodeAct agent (whose resulting end-to-end system is referred to as \osa-v2). 

\subsubsection{\osa-v1 with ReAct Agent} 
\label{sec:ODS-v1}

Our first Open Reasoning Agent is based on Chain-of-thought (CoT) reasoning \cite{kojima2022zero,wei2022chain} and ReAct \cite{yao2022react} agent.

{\bf Chain-of-Thought (CoT)} prompting elicits impressive reasoning capabilities in LLM agents by encouraging the model to stop and think prior to answering \cite{wang2024survey,chu2023survey}.  Zero-shot CoT consists of simply appending the phrase ``Let's think step by step." to the end of the prompt fed to the model. \cite{kojima2022zero} The results are even further improved by combining CoT with the few-shot prompting, an approach we adopt. Few-shot CoT consists of appending several CoT examples to the prompt as a form of in-context learning \cite{wei2022chain}.

{\bf Chain-of-Thought Self-Consistency (CoT-SC)} further improves on the greedy decoding techniques utilized in simple CoT. Instead of naively taking a single greedy reasoning path, CoT-SC samples multiple different pathways and compares them to then select the most consistent answer among all pathways. This has led to signficiant improvements upon naive CoT in multiple reasoning tasks including arithmetic and question-answering tasks. \cite{wang2022self}

\medskip
{\bf ReAct Framework.}  ReAct (Reasoning and Action) is a language model framework that synergistically combines reasoning steps with action execution to enhance task completion and decision-making capabilities \cite{yao2022react}.
ReAct combines reasoning and action through an iterative process that enables language models to perform complex tasks with enhanced reliability. The framework consists of three interleaved components:
\begin{equation*}
    P(a_t|s_t) = f(r_t, o_{t-1}, h_{t-1})
\end{equation*}
where $a_t$ is the current action, $s_t$ the current state, $r_t$ the reasoning output, $o_{t-1}$ the previous observation, and $h_{t-1}$ the interaction history.

The framework implements tool integration through a standardized interface:
\begin{verbatim}
Thought: [Reasoning trace]
Action: Tool[parameters]
Observation: [Result]
\end{verbatim}

 Few-shot learning in ReAct leverages a small set of demonstrative examples to guide the model's reasoning and action patterns.

\begin{figure}[h!]
        \centering
        \begin{lstlisting}[basicstyle=\ttfamily\tiny, frame=single]
<Question>
What is the capital of France?
</Question>
<Thought>
I need to determine the capital city of France.
</Thought>
<Action>
search_internet
</Action>
<Action_Input>
"capital of France"
</Action_Input>
<Observation>
The capital of France is Paris.
</Observation>
<Final_Answer>
Paris
</Final_Answer>
        \end{lstlisting}
        \caption{Illustration of the ReAct prompt structure utilized in \osa-v1.}
    \label{fig:react_prompt}
\end{figure}

\medskip\noindent
{\bf Dynamic Few-Shot Learning.} 
\noindent ReAct implements few-shot learning through example-based prompting, where a small set of demonstrations guides the model's reasoning and action patterns. To optimize prompt efficiency, a dynamic few-shot selection system utilizes vector similarity matching to retrieve the most relevant examples for each task, maintaining performance while reducing prompt complexity.

We ran a community campaign to design the 200 ReAct prompts used in our few-shot template. Participants were asked to apply their own  intuitions of reasoning to the prompt designs, resulting in a wide range of approaches. They were provided with a template describing the structure of ReAct prompts (Thought/Action/Action Input/Observation) and some sample queries from a test set. We made sure not to grant them access to the benchmarks themselves. The campaign substantially improved the performance of our ReAct agent due to the diverse range of thinking processes represented in the ReAct prompts.  Examples from the resulting few-shot prompts are provided in Appendix~\ref{app:prompts}. 

\medskip\noindent
{\bf Tool Integration.}  The framework is integrated with external tools, enabling more sophisticated problem-solving capabilities. We use three tools in \osa-v1: 

\begin{enumerate}
    \item \textbf{Web Search:} Open Search Tool from Section~\ref{sec:OpenSearchTool}. 
    \item \textbf{Mathematical Processing:} integration with Wolfram Alpha API for handling arithmetic and complex mathematical computations.
    \item \textbf{Continue Thinking:} continued reasoning to break down complex queries using the base LLM.
\end{enumerate}



\medskip\noindent{\bf \osa-v1.} We integrate CoT consistency sampling, the ReAct agentic framework, and few-shot prompting in Open Reasoning Agent of \osa-v1. 

For any query, we first run Open Search Tool and the resulting context, along with the original query, is fed into the ReAct Agent. The ReAct agent uses prompts that consist of structured steps: $<Thought>, <Action>, <Action\_Input>,$ and $<Observation>$, beginning with a $<Question>$ and concluding with a $<Final\_Answer>$. These steps iterate until a response is returned, though the agent may not return an answer if insufficient information is available. 

The ReAct agent has three action options. ``Continue thinking" ($<Action> = continue\_think$) extends the reasoning process for complex problem decomposition. ``Search" ($<Action> = search\_internet$) leverages OpenPerplex for finding factual information like dates, names, and academic content.  ``Calculate" ($<Action> = calculate$) connects to the Wolfram Alpha API for handling numerical computations that foundation models typically struggle with. 

When the ReAct agent fails to provide an answer—as determined by a Llama-3-8B judge model—the system defaults to Chain-of-Thought Self-Consistency. This backup process makes $r$ calls to the LLM, clusters similar responses, and returns a random response from the largest cluster, ultimately leading to the final answer.

In the following example taken from the FRAMES benchmark, we can observe the reasoning advantage \osa-v1 has over state-of-the-art closed models such as Perplexity's Sonar Reasoning Pro. Both \osasp and the Sonar Reasoning Pro model get confused between potential answers. However, \osa correctly identifies 112 inches as the correct answer and runs an additional check using the Wolfram-Alpha tool to convert the answer to 2,845mm as required. Conversely, Sonar Reasoning Pro just responds with the incorrect answer of 2,858mm. 

\begin{figure}[h!]
    \centering
    \begin{lstlisting}[basicstyle=\ttfamily\tiny, frame=single]
A question from FRAMES: 
    How old would the 1975 winner of the Lenore Marshall Poetry Prize have been if they were still alive on the date when Rupi Kaur released her book titled, "Milk and Honey"?
Correct Answer: 
    90 years
Perplexity response:
    To determine the age of the 1975 winner of the Lenore Marshall Poetry Prize if they were still alive when Rupi Kaur released her book "Milk and Honey" in 2014, we need to consider the following: 
        1. **Year of the Prize**: The Lenore Marshall Poetry Prize was awarded in 1975. 
        2. **Release Year of "Milk and Honey"**: Rupi Kaur released "Milk and Honey" in 2014. 
        3. **Age Calculation**: To find the age, subtract the year of the prize from the year of the book release. 
    If the winner was alive in 2014, we calculate their age as follows: \[ \text{Age in 2014} = 2014 - 1975 = 39 \text{ years} \] However, this calculation only gives the difference in years between the two events. To find the actual age of the person, we need to know their birth year. Since the specific winner's birth year is not provided, we can only state that if the 1975 winner was still alive in 2014, they would be 39 years older than they were in 1975. For example, if the winner was 40 years old in 1975, they would be \(40 + 39 = 79\) years old in 2014. Without the exact birth year or age at the time of the award, we cannot provide a precise age[1][4]
ODS-v1+Llama3.1-70B response: 
    '$Initial\_Thought$''
        To answer this question, I\'ll need to follow these steps:
            1. Identify the winner of the 1975 Lenore Marshall Poetry Prize.
            2. Determine the birth year of the winner.
            3. Calculate the age of the winner as of the publication date of Rupi Kaur\'s book "Milk and Honey" (November 4, 2014).
        Let\'s start by identifying the winner of the 1975 Lenore Marshall Poetry Prize. According to the provided context, the winner was Cid Corman., 
    {'Thought': 'Now that I have the birth year of Cid Corman, I can calculate his age as of the publication date of Rupi Kaur\'s book "Milk and Honey"(November 4, 2014)'},  
        Do I need further steps? No.', 
    {'Action': calculate}, {'Action\_Input': '2014-1924' 
    'Thought': 'final\_result': 'Cid Corman was born in 1924, so as of the publication date of Rupi Kaur\'s book "Milk and Honey" (November 4, 2014), he would have been 90 years old. Do I need further steps? No}, 
    {'Action': 'Done'}'
    \end{lstlisting}
    \caption{An example question and answer from FRAMES where \osa-v1 uses the Wolfram calculator tool to accurately calculate the age difference, resulting in the correct answer of 90. Conversely, Perplexity pursues the wrong reasoning path to report 79 years of age.}
    \label{fig:example_calculate1}
\end{figure}

\subsubsection{\osa-v2 with CodeAct Agent} 
\label{sec:ODS-v2}

\textbf{Chain of Code (CoC).} While Chain-of-Thought (CoT) has proven effective in semantic reasoning, it often encounters challenges with tasks requiring precise numeric or symbolic computations. To address these limitations, Chain-of-Code \cite{li2024chain} (CoC) leverages the code-writing capabilities of LLMs to generate and execute code, or pseudocode, to tackle both algorithmic and semantic problems. This approach not only broadens the scope of reasoning questions LMs can address but also enhances their accuracy in solving complex tasks. CoC outperforms traditional CoT methods across various benchmarks, highlighting the potential of integrating code generation and execution within LLMs to achieve more robust reasoning capabilities.

\medskip\noindent

\textbf{CodeAct.}
Recent advancements \cite{lv2024codeact} have shown that generating executable Python code for tool calling yields a significant boost in performance compared to conventional JSON-based approaches. In particular, LLM's are inherently adept at compressing the action-space of tasks using code. Additionally, code, as a mode of representation, is more naturally suited for action taking as it can be composed, modularized, and generalized much easier than JSON based approaches.

\medskip\noindent

\begin{figure}[h]
    \centering
    \includegraphics[width=0.75\linewidth]{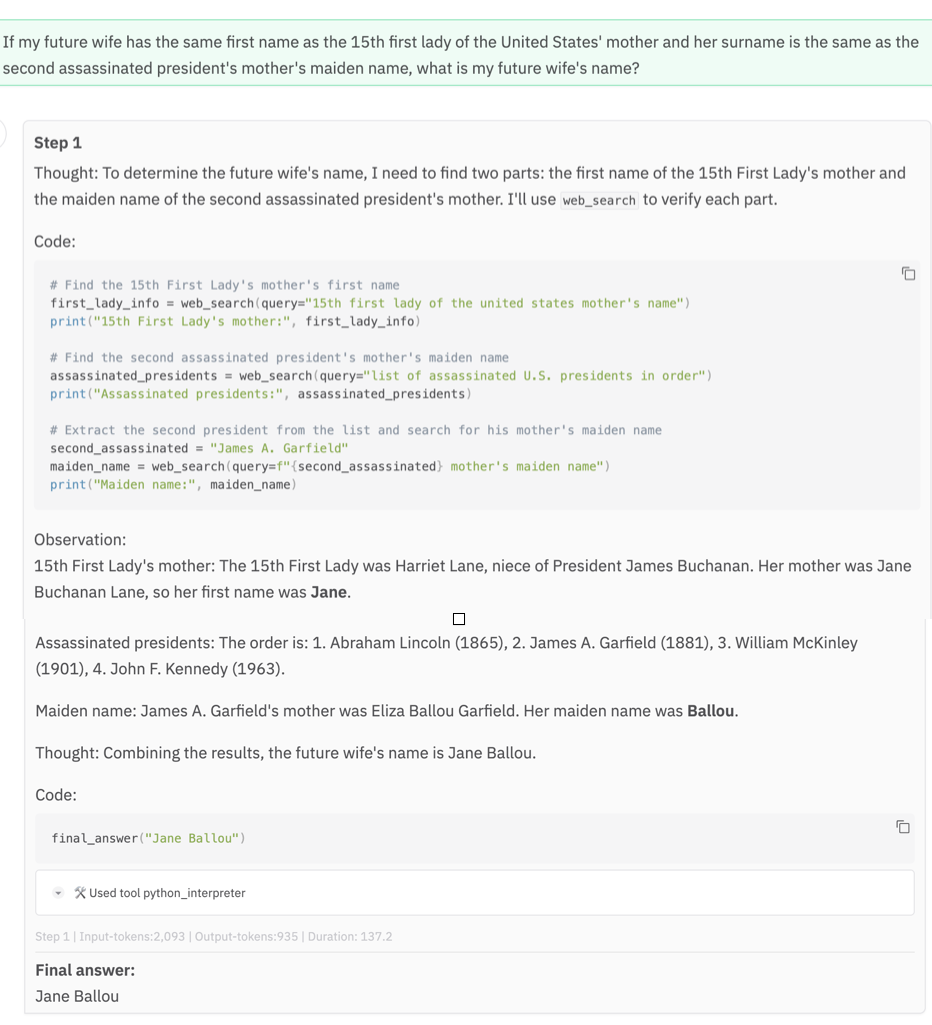}
    \caption{CodeAct agent in ODS-v2 answering a multi-hop question.}
    \label{fig:codeagent}
\end{figure}

In \textbf{ODS-v2}, we adapt our search tool to work with SmolAgents's \cite{smolagents} framework as it allows for customization and can be easily distributed. Our most basic reasoning agent powered by search uses \cite{smolagents} CodeAgent (A variation of CodeAct) with access to our search tool as shown in \ref{fig:codeagent}. More advanced iterations of \textbf{ODS-v2} involve multiple tools and agents working in-concert to solve more complicated tasks that may, or may not, involve search.

\medskip\noindent

\section{Experiments} 
 \label{sec:results}

{\bf Baselines.} We compare against the popular closed-source search AIs from Perplexity \cite{perplexity2024} (their default search AI, which we call Perplexity, and an advanced reasoning search AI, which is called Perplexity Sonar Reasoning Pro \cite{perplexity_sonar_pro}) and the state-of-the-art search AI from OpenAI: GPT-4o Search Preview \cite{gptsearch}.  These are the state-of-the-art AI solutions with access to search engines. As separate baselines, we also compare against large language models: GPT-4o, Llama-3.1-70B, and DeepSeek-R1. Although these models do not have access to the Internet, we demonstrate in Table~\ref{tab:main} that the LLMs with reasoning capabilities are surprisingly good at our evaluation benchmarks of FRAMES  \cite{krishna2024} and SimpleQA \cite{wei2024measuring}.


\subsection{Numerical analyses on two evaluation benchmarks: FRAMES and SimpleQA}

We use two evaluation benchmarks: 
  FRAMES \cite{krishna2024} and SimpleQA \cite{wei2024measuring}. Originally, SimpleQA is intended to test frontier-models on factuality without web-browsing, and   FRAMES is intended to test factuality and retrieval of models on single to multi-hop queries where the ground truth Wikipedia articles are given. In our scenario, we use both of them to evaluate the accuracy of search AIs with access to the Internet.

\subsubsection{Numerical analysis on FRAMES} 

FRAMES (Factuality, Retrieval, And
reasoning MEasurement Set)  dataset introduced in  \cite{krishna2024} comprises
824 challenging multi-hop questions requiring
integrating multiple
sources from Wikipedia. The best single-query search approach reported in \cite{krishna2024}  achieves a score of 47.4\% with Gemini-Pro-1.5-0514 (5/14/2024 release) when using 4 retrieved documents from a single query that have the highest
BM25 score \cite{robertson1995okapi} from 
a Wikipedia data dump\footnote{https://www.tensorflow.org/datasets/catalog/wikipedia}. \osa-v1+DeepSeek-R1 achieves 56.7\%  under the same condition of using a single web search per query in Table~\ref{tab:ablation_prompt}. This is a significant improvement over the state-of-the-art reasoning LLMs (such as DeepSeek-R1 and GPT-4o) and search AIs (such as Perplexity and Perplexity Sonar Reasoning Pro) as shown in Table~\ref{tab:main}.
The CodeAct-based agent of \osa-v2+DeepSeek-R1 chooses to use more searches, thus achieving 75.3\% on FRAMES using 3.39 searches  per query on average.

With our reasoning agent, \osa-v1+DeepSeek-R1 has the option to run more searches but chooses not to. On the other hand, when \osa-v1 uses Llama3.1-70B as a base model, it uses more web searches to improve accuracy, as shown in the table and also illustrated with an example in Figure~\ref{fig:example_search1}. 

\begin{table}[h!]
\begin{center}
\begin{tabular}{|c|cc|cc|}
\hline 
 & \multicolumn{2}{c|}{Accuracy (\%)} & \multicolumn{2}{c|}{\# of web searches per query} \\
\osa+base-model&SimpleQA&FRAMES&SimpleQA&FRAMES\\ \hline
\osa-v1+Llama3.1-70B & 83.4 & 49.5 & 1.09 & 1.05 \\\hline 
\osa-v1+DeepSeek-R1 & 87.7 & 56.7 & 1.00  & 1.00  \\\hline
\osa-v2+DeepSeek-R1 & 88.3  & 75.3  & 1.45  & 3.39   \\\hline

\end{tabular}
\end{center}
\caption{How many times \osasp  searches the web, on average per query, for the two benchmarks. \osasp adapts to the quality of the first search result and the model output to judiciously use the extra search only when deemed necessary.}
\label{tab:ablation_prompt}
\end{table}

It has been reported in \cite{krishna2024} that the accuracy on FRAMES can be improved with multiple web searches. For example, performance similar to \osa-v1+DeepSeek-R1 can be achieved by Gemini-Pro-1.5-0514 by running the search engine 15 times with rephrased prompts. However, such a multi-step multi-search approach is inefficient since it calls a fixed number, for example 15, of searches regardless of the hardness of the example or the correctness of the model output. This is one of the motivations for our agentic framework, where our reasoning agent adaptively selects whether to search again or not based on the initial search quality and the model output. This is shown Table~\ref{tab:ablation_prompt}, where \osa-v1 uses more web searches when paired with a weaker base model of Llama3.1. The example in Figure~\ref{fig:example_search1} shows how \osa-v1 uses the search judiciously when the initial search is insufficient.  Similarly for \osa-v2, it chooses to use fewer web searches on SimpleQA, which is a relatively easier task requiring a single correct source of information, compared to FRAMES, which is a much more  complex task requiring multiple sources of information. 

\begin{figure}
\centering
\includegraphics[width=.45\textwidth]{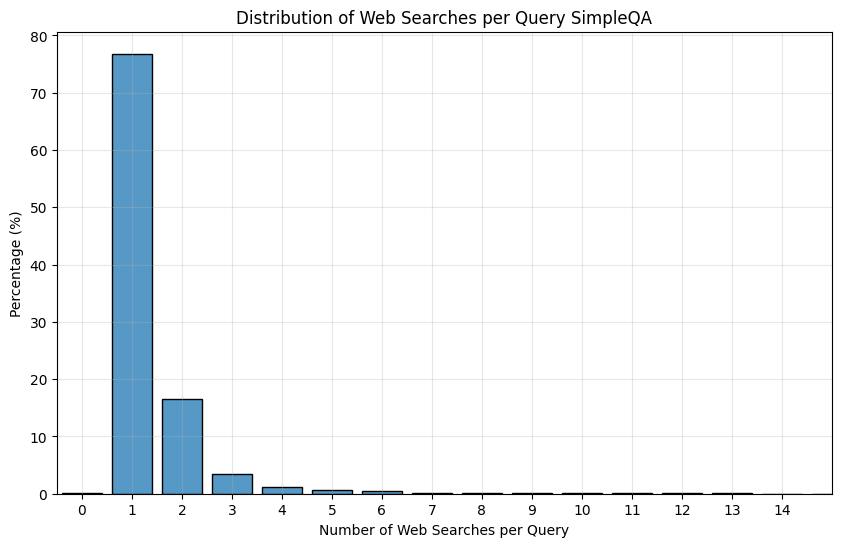}
\includegraphics[width=.45\textwidth]{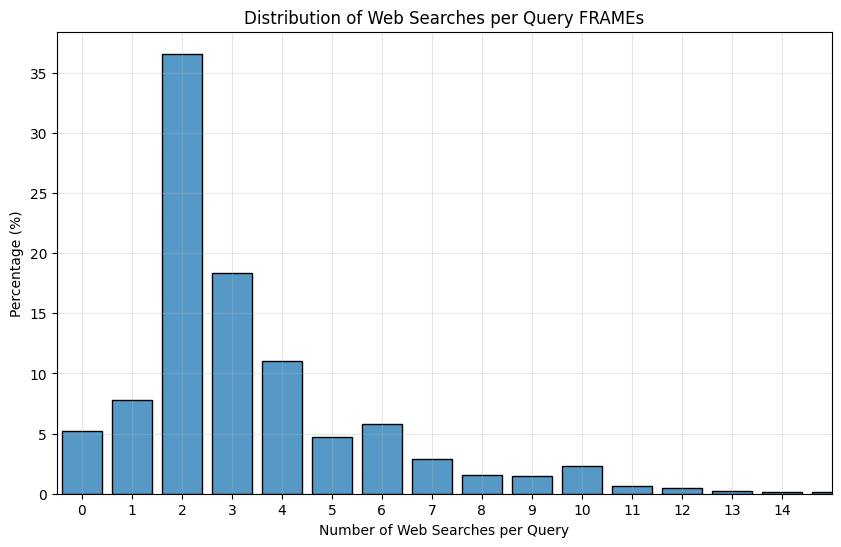}
    \caption{The histogram of how many searches are run by the CodeAct-based \osa-v2 reveals that it chooses to use less number of web searches on SimpleQA, which is relatively easier, compared to FRAMES, which is much more  complex. }
    \label{fig:search_frequency}
\end{figure}

\begin{figure}[h!]
        \centering
        \begin{lstlisting}[basicstyle=\ttfamily\tiny, frame=single]
A question from FRAMES: 
    What year was the band leader of the group who originally performed the song sampled in Kanye West's song Power born?
Correct Answer: 
    1946
Perplexity response: 
    The song "Power" by Kanye West samples several tracks, including "21st Century Schizoid Man" by King Crimson.
    The lead vocalist and bassist of King Crimson during the time of that song was Greg Lake.
    Greg Lake, the band leader and original lead vocalist of King Crimson for the song "21st Century Schizoid Man," was born on November 10, 1947[2][4]. 
ODS-v1+Llama3.1-70B response: 
    '$Initial\_Thought$': 
        To answer this question, I\'ll need to follow these steps:
        1. Identify the song sampled in Kanye West\'s song "Power".
        2. Identify the band that originally performed the sampled song.
        3. Identify the band leader of the original band.
        4. Find the birth year of the band leader.
        Based on the context, I can start by identifying the song sampled in Kanye West\'s song "Power". According to the context, the song sampled is "21st Century Schizoid Man" by King Crimson.    
        Next, I can identify the band that originally performed the sampled song, which is King Crimson.
        Now, I need to identify the band leader of King Crimson. According to the context, the founder of King Crimson is Emilio Castillo, but this is not correct. After re-reading the context, I found that the founder of King Crimson is not explicitly mentioned, but it is mentioned that the band was born in October 1968.
        However, I can infer that the band leader of King Crimson is likely Robert Fripp, who is a well-known guitarist and composer of the band. But I couldn\'t find any information about Robert Fripp\'s birth year in the given context.
        Let me continue searching for the birth year of the band leader of King Crimson...  
    'Action': '$search\_internet$', '$Action\_Input$': 'Robert Fripp birth year'
    'Thought': 
        Based on the search results, I found that Robert Fripp, the band leader of King Crimson, was born on May 16, 1946.', 
    '$Action$': 'Done' 

   
        \end{lstlisting}
        \caption{An example question and answer from FRAMES where \osa-v1+Llama3.1-70B realizes that a second search is necessary, searches again to find the birth year of the lead singer of King Crimson, and correctly answers the question with “1946”. Perplexity \cite{perplexity2024} on the other hand could not figure out the leader of the band King Crimson.}
    \label{fig:example_search1}
\end{figure}

\subsubsection{Numerical analyses on SimpleQA} 
SimpleQA, introduced in \cite{wei2024measuring}, is a  benchmark evaluation dataset, adversarially
collected against GPT-4, and the questions are designed so that there is only a single, indisputable
answer. It consists of 4326 short-form factuality questions, e.g., ``On which U.S. TV station did the Canadian reality series *To
Serve and Protect* debut?'' or ``Who received the IEEE Frank Rosenblatt Award in 2010?''. An LLM-as-a-judge with GPT-4o-mini is used to check for correctness of the response w.r.t.~the ground truths, e.g., ``KVOS-TV'' or ``Michio Sugeno''.  
It consists of questions from diverse topics including Science \& Technology (n=858), Politics (n=709),
and Art (n=550), Geography (n=424), 
Sports (n=368), 
Music (n=341), 
TV shows (n=293), 
History (n=173), and 
Video games (n=135). It is reported in \cite{wei2024measuring} that GPT-4o scores 38.2\% accuracy, Claude-3.5-sonnet (2024-06-20 release) gets 28.9\%, and OpenAI o1-preview gets 42.7\%. \osa-v1 and \osa-v2 improve upon these closed-source Perplexity solutions, achieving 87.7\% and 88.3\%, respectively (Table~\ref{tab:main}), by leveraging the high quality search results provided by our Open Search Tool (Section~\ref{sec:OpenSearchTool}) together with the Open Reasoning Agent framework (Section~\ref{sec:OpenReasoningAgent}). It is important to note that the gain comes from both a high quality search tool (as shown in an example in Figure~\ref{fig:example_simpleqa_searchtool}) and the reasoning agent (as shown in an example in Figure~\ref{fig:example_simpleqa_calculator}). 

\begin{figure}[h!]
    \centering
    \begin{lstlisting}[basicstyle=\ttfamily\tiny, frame=single]
A question from SimpleQA: 
    What is the name of track 10 on disc 2 of the Sen no Kiseki IV - The End of Saga - original soundtrack?
Correct Answer: 
    Break In 
Perplexity Sonar Reasoning Pro response:
    Okay, lets tackle this query. The user is asking for the name of track 10 on disc 2 of the Sen no Kiseki IV - The End of Saga original soundtrack. First, I need to check the search results provided. Let me go through each source to find the track listing for disc 2. Looking at result [1], which is from Apple Music, it lists the tracks but only shows the first few tracks of disc 1. The full track list is not visible here, so this might not help directly. Result [2] from Spotify mentions some track names like "Beyond Ten Millions of Nights" and "Path to the..." but does not list track numbers or disc divisions clearly. Not enough info here. Result [3] .....But in the provided data from result [7], I do not see a breakdown by disc. Wait, maybe result [7] includes all tracks as a single list? Let me check again. Result [7] lists "Number of Files: 69" and "Total Filesize: 365 MB (MP3), 3,866 MB (FLAC)." But the actual track list is not shown here. The user needs to know which specific track is number 10 on disc 2. Without seeing the full track list split by discs, this might not help directly. Result [8] on SoundCloud has a track titled which translates to "Each Persons Resolution," but it is part of disc 3 according to the URL. So that is not relevant here. If we consider that each disc has around ~23 tracks, then Disc 2 Track 10 would be overall Track 33 or so, but without knowing exact names, this approach will not work. Alternatively, maybe someone can infer based on other games soundtracks structure
ODS-v1+DeepSeek-R1 response: 
    Initial_Thought: <think>
        Okay, let's try to figure out the name of track 10 on disc 2 of the Sen no Kiseki IV original soundtrack. First, I need to go through the provided context and look for any mentions of disc 2 track listings.
        Looking at the context snippets, there are several mentions of disc 2. One of them says: "Disc 2 ; 2.10, Break In Break In, C : HS ; 2.11, Trump Card for the Worst Situation , C : HS ; 2.12, Bold Assailants..." ....So putting this together, track 10 on disc 2 is likely "Break In". </think>
    The name of track 10 on disc 2 of Sen no Kiseki IV -THE END OF SAGA- Original Soundtrack is "Break In". This is confirmed by context snippets explicitly listing track 2.10 as "Break In Break In".
    Action: Done
    final_result: The name of track 10 on disc 2 of the Sen no Kiseki IV - The End of Saga - original soundtrack is "Break In"
    \end{lstlisting}
    \caption{An example question and answer from SimpleQA where the \osa-v1  leverages the high quality retrieved context from Open Search Tool to identify the correct answer, by cross-checking  multiple sources. Perplexity Sonar Reasoning Pro \cite{perplexity_sonar_pro} failed to retrieve the relevant information in the search. }
    \label{fig:example_simpleqa_searchtool}
\end{figure}

\begin{figure}[h!]
    \centering
    \begin{lstlisting}[basicstyle=\ttfamily\tiny, frame=single]
A question from SimpleQA: 
    The Jensen Interceptor (1950), produced from 1950 to 1957, had a wheelbase measurement of what in millimeters?
Correct Answer: 
    2,845 mm 
Perplexity Sonar Reasoning Pro response:
    Let's tackle this query about the Jensen Interceptor (1950) wheelbase in millimeters. The user is asking specifically for the wheelbase measurement of the model produced between 1950 and 1957. First, I'll look through the provided search results to find mentions of the wheelbase. Going through each source: - Result [1] from automobile-catalog.com states: "Wheelbase:2858 mm / 112.5 in". That seems directly relevant. - Result [3] from motor-car.net lists "Wheelbase: 112 in (2,845 mm)". The Jensen Interceptor produced from 1950 to 1957 had a wheelbase of 2,858 mm (112.5 inches)[1][8]. This measurement is consistent across technical specifications from multiple sources detailing the first-generation Interceptor's dimensions[1][3][8].

ODS-v1+DeepSeek-R1 response: 
     'Initial_Thought': '<think>
         Okay, let\'s tackle this question about the Jensen Interceptor (1950) wheelbase in millimeters. The user is asking specifically about the model produced from 1950 to 1957. First, I need to go through the provided context and find any mentions of the wheelbase measurement for that specific model and years.
         Looking at the context snippets, there are several mentions of wheelbase measurements. Let me list out the relevant ones:
         1. "Jensen Interceptor (1950) ; 4-speed manual, 112 in (2,845 mm), 183 in (4,648 mm)." Here, 112 inches is converted to 2,845 mm. This seems to directly mention the 1950 model's wheelbase.
         2. Another snippet says: "Wheelbase: 112 in (2,845 mm). Length, 183 in (4,648 mm). Width, 66 in (1,676 mm)." ...}, 
    {'Thought': "The context mentions both 112 inches (2844.8 mm) and 112.5 inches (2857.5 mm). The 1950 model's original specs from the article state 112 inches (2845 mm), while later references to the cabriolet and other variants might use 112.5 inches. Since the user specified the 1950-1957 model, the primary value from the 1950 article is 112 inches. Let me convert this value."}, 
    {'Action': 'calculate'}, {'Action_Input': '112 inches to millimeters'}, 
    {'Observation': {'2845mm"'}}, {'Thought': ''}, {'Action': 'Done'},
    'final_result': 'The wheelbase measurement of the Jensen Interceptor (1950) produced from 1950 to 1957 is 2,845 millimeters.'' 
    \end{lstlisting}
    \caption{An example question and answer from SimpleQA where the Open Reasoning Agent correctly identifies 112 inches as the correct answer and makes an additional check with the Wolfram-Alpha API to convert the answer to 2,845mm. Conversely, Perplexity Sonar Reasoning Pro \cite{perplexity_sonar_pro} gets confused between 112.5 inches and 112 inches as potential answers. We abbreviated the responses here and refer to Appendix~\ref{app:examples} for the complete responses.}
    \label{fig:example_simpleqa_calculator}
\end{figure}

In Table~\ref{tab:ablation_simpleqa}, we survey SimpleQA accuracy results reported in other sources together with \osa-v1+Llama3.1-70B, \osa-v1+DeepSeek-R1, and \osa-v2+DeepSeek-R1. With the help of the reasoning power of DeepSeek-R1, \osa-v1+DeepSeek-R1 improves over  \osa-v1+Llama3.1-70B. There are more recent closed-source solutions such as GPT-4o Search Preview \cite{gptsearch}, Exa \cite{exa}, Linkup \cite{linkup}, and Perplexity Deep Research \cite{perplexity_deep_research} that claim to achieve higher SimpleQA scores. \osasp provides an open-source solution that can significantly close the gap between the open search AI solutions and the closed ones.

\begin{table}[h!]
\begin{center}
\begin{tabular}{|c|c|}
\hline 
Search AI & SimpleQA (\%)\\\hline 
\multicolumn{2}{|c|}{{\em
Closed-source}} \\\hline

o1-mini \cite{simple_evals} & $7.6$ \\
GPT-4o-mini-2024-07-18 \cite{simple_evals} & $9.5$ \\
o3-mini-low \cite{simple_evals} & $13.0$ \\
o3-mini \cite{simple_evals} & $13.4$ \\
o3-mini-high \cite{simple_evals} & $13.8$ \\
Grok 3 Mini Beta  \cite{grok3}  & $21.7$ \\
Claude 3 Opus \cite{simple_evals} & $23.5$ \\
GPT-4-turbo-2024-04-09 \cite{simple_evals} & $24.2$ \\
Claude 3.5 Sonnet \cite{simple_evals} & $28.9$ \\
GPT-4o \cite{grok3} & $38.2$ \\
GPT-4o-2024-11-20 \cite{simple_evals} & $38.8$ \\
GPT-4o-2024-05-13 \cite{simple_evals} & $39.0$ \\
GPT-4o-2024-08-06 \cite{simple_evals} & $40.1$ \\
o1 \cite{simple_evals} & $42.6$ \\
o1-preview \cite{simple_evals} & $42.7$ \\
Grok 3 Beta \cite{grok3} & $43.6$ \\
Gemini 2.0 Pro \cite{grok3} & $44.3$ \\
Perplexity Sonar \cite{perplexity_sonar_pro} & $77.3$ \\
Perplexity~\cite{perplexity2024} & 82.4\\
Perplexity Sonar Reasoning Pro \cite{perplexity_sonar_pro} & $85.8$ \\
GPT-4o Search Preview \cite{gptsearch} & $90.0$ \\ 
Exa \cite{exa} & $90.0$ \\ 
Linkup \cite{linkup} & $90.1$ \\
Perplexity Deep Research \cite{perplexity_deep_research}& $\textbf{93.9}$ \\
\hline
\multicolumn{2}{|c|}{{\em
Open-source}} \\\hline
Qwen 2.5 \cite{deepseek} & $9.1$ \\
Llama3.1-70B& $20.4^*$ \\
DeepSeek-V3 \cite{grok3} & $24.9$ \\
DeepSeek-R1 \cite{guo2025deepseek} & $82.4$ \\
\osa-v1+Llama3.1-70B  & $84.0^*$ \\
\osa-v1+DeepSeek-R1  & $87.7^*$ \\
\osa-v2+DeepSeek-R1  & $\textbf{88.3}^*$ \\
\hline
\end{tabular}
\end{center}
\caption{\osa-v2 achieves the best accuracy on SimpleQA among open-source solutions. $^*$ indicates evaluations run by us. Otherwise, the SimpleQA accuracy is taken from the cited source in each row. }
\label{tab:ablation_simpleqa}
\end{table}

\subsection{Ablation study of \osa}



In Table~\ref{tab:ablation}, we run an ablation study for various parts of the ReAct-based Open Reasoning Agent for \osa-v1. Llama3.1-70B alone achieves 21.2\% on a subset of 500 randomly selected examples from SimpleQA (we use a subsampled evaluation dataset for efficiency, just in this section) and 34.3\% on FRAMES. With our Open Search Tool, which we abbreviated as Search in the table, SimpleQA increases dramatically to 82.4\% but FRAMES suffers a performance drop to 27.6\%. Our Open Reasoning Agent has two parts: Chain-of-Thought ReAct with self-consistency decoding (CoT-ReAct) and Few-shot prompting (FewShot). Adding these components one-by-one, we get gradual improvements in both subsampled SimpleQA and FRAMES. This shows that each component of \osa-v1 contributed to the performance achieved. Finally, by replacing the base-model from Llama3.1-70B with a more powerful reasoning model of DeepSeek-R1, we show that \osa-v1 leverages the reasoning capabilities of the LLM to achieve even better performance.


\begin{table}[h!]
\centering
\renewcommand{\arraystretch}{1.2}
\begin{tabular}{l|cc}
\hline
Search AI Solution & 500 subset of  & FRAMES  \\
 & SimpleQA (\%)&  (\%)\\
\hline
Llama3.1-70B & 21.2 & 34.3 \\
OpenSearchTool + Llama3.1-70B &82.4 & 27.6 \\
OpenSearchTool + Llama3.1-70B + CoT-ReAct &  87.2 & 37.4 \\
OpenSearchTool + Llama3.1-70B + CoT-ReAct + FewShot & 87.8 & 49.5 \\ \hline
OpenSearchTool + DeepSeek-R1 + CoT-ReAct + FewShot  & 90.4 &  56.7 \\
\end{tabular}
\vspace{0.5cm} 
\caption{Ablation study of Open Search Tool, Llama3.1-70B/DeepSeek-R1, and  the two components of the ReAct-based Open Reasoning Agent in  \osa-v1: Chain-of-Thought ReAct with self-consistency decoding (CoT-ReAct) and Few-shot prompting (FewShot). The SimpleQA accuracy is only reported on the 500 randomly sampled examples, for efficiency. Each component is critical in achieving the state-of-the-art performance.}
\label{tab:ablation}
\end{table}






\section{Conclusion}
\label{sec:conclusion}



To bridge the increasingly widening gap between state-of-the-art proprietary search AI solutions such as Perplexity Sonar Reasoning Pro \cite{perplexity_sonar_pro} and GPT-4o Search Preview \cite{gptsearch} and their open-source alternatives \cite{vu2023freshllmsrefreshinglargelanguage,openperplex,perplexica}, we introduce Open Deep Search (ODS). This open-source search AI can be seamlessly combined with any LLM of the user's choice in a plug-and-play style. This allows ODS to harness the latest advances in reasoning LLMs and achieve increasingly accurate performance. When using DeepSeek-R1, ODS can achieve 75.3\% accuracy on FRAMES benchmark, surpassing GPT-4o Search Preview released on 3/11/2025 by 10\% (Table~\ref{tab:main}). On another benchmark of SimpleQA, ODS significantly narrows the gap between open-source and closed-source solutions. This significant milestone is achieved by the synergy between the two components of ODS: Open Search Tool and Open Reasoning Agent. 

\begin{itemize} 
\item We introduce {\bf Open Search Tool} to provide high quality retrievals from the Internet, to be used as a tool by the reasoning agentic framework within ODS. 
\item {\bf Open Reasoning Agent} interprets the given task and completes it by calling available tools, ranging from the search tool (our proposed Open Search Tool), calculator (provided by Wolfram Alpha API), reflection (using the provided base LLM), and code interpreter (we use Python interpreter). 
\end{itemize} 
Together with the latest powerful reasoning LLMs, these two open-source components that make up ODS ensure that we achieve the state-of-the-art performance on search. 
The gain of Open Search Tool over the proprietary counterparts is demonstrated in, for example, Figure~\ref{fig:example_simpleqa_searchtool} and other examples in Appendix~\ref{app:examples}. The gain of Open Reasoning Agent is demonstrated in, for example, Figures~\ref{fig:example_search1} and \ref{fig:example_simpleqa_calculator}, and other examples in Appendix~\ref{app:examples}. 

We provide two versions of Open Reasoning Agent: one based on ReAct and the other based on CodeAct. We publicly release all open-source implementations and invite the open-source community to build upon our work and further innovate, starting from our state-of-the-art search AI solution.




\newpage
\bibliographystyle{plain}
\bibliography{sample}  

\begin{thebibliography}{10}

\bibitem{exa}
Will Bryk.
\newblock https://exa.ai/blog/api-evals, 2025.

\bibitem{chen2024mindsearch}
Zehui Chen, Kuikun Liu, Qiuchen Wang, Jiangning Liu, Wenwei Zhang, Kai Chen,
  and Feng Zhao.
\newblock Mindsearch: Mimicking human minds elicits deep ai searcher.
\newblock {\em arXiv preprint arXiv:2407.20183}, 2024.

\bibitem{chu2023survey}
Zheng Chu, Jingchang Chen, Qianglong Chen, Weijiang Yu, Tao He, Haotian Wang,
  Weihua Peng, Ming Liu, Bing Qin, and Ting Liu.
\newblock A survey of chain of thought reasoning: Advances, frontiers and
  future.
\newblock {\em arXiv preprint arXiv:2309.15402}, 2023.

\bibitem{deepseek}
DeepSeek.
\newblock https://www.deepseek.com, 2025.

\bibitem{guo2025deepseek}
Daya Guo, Dejian Yang, Haowei Zhang, Junxiao Song, Ruoyu Zhang, Runxin Xu,
  Qihao Zhu, Shirong Ma, Peiyi Wang, Xiao Bi, et~al.
\newblock Deepseek-r1: Incentivizing reasoning capability in llms via
  reinforcement learning.
\newblock {\em arXiv preprint arXiv:2501.12948}, 2025.

\bibitem{kojima2022zero}
T.~Kojima, S.~S. Gu, M.~Reid, Y.~Matsuo, and Y.~Iwasawa.
\newblock Large language models are zero-shot reasoners.
\newblock {\em Advances in Neural Information Processing Systems},
  35:22199--22213, 2022.

\bibitem{kovsprdic2024verif}
Milo{\v{s}} Ko{\v{s}}prdi{\'c}, Adela Ljaji{\'c}, Bojana Ba{\v{s}}aragin,
  Darija Medvecki, and Nikola Milo{\v{s}}evi{\'c}.
\newblock Verif. ai: Towards an open-source scientific generative
  question-answering system with referenced and verifiable answers.
\newblock {\em arXiv preprint arXiv:2402.18589}, 2024.

\bibitem{krishna2024}
Satyapriya Krishna, Kalpesh Krishna, Anhad Mohananey, Steven Schwarcz, Adam
  Stambler, Shyam Upadhyay, and Manaal Faruqui.
\newblock Fact, fetch, and reason: A unified evaluation of retrieval-augmented
  generation.
\newblock 2024.

\bibitem{lewis2020retrieval}
Patrick Lewis, Ethan Perez, Aleksandra Piktus, Fabio Petroni, Vladimir
  Karpukhin, Naman Goyal, Douwe Kiela, et~al.
\newblock Retrieval-augmented generation for knowledge-intensive nlp tasks.
\newblock {\em Advances in Neural Information Processing Systems},
  33:9459--9474, 2020.

\bibitem{li2024generative}
Alice Li and Luanne Sinnamon.
\newblock Generative ai search engines as arbiters of public knowledge: An
  audit of bias and authority.
\newblock {\em Proceedings of the Association for Information Science and
  Technology}, 61(1):205--217, 2024.

\bibitem{li2024chain}
Chengshu Li, Jacky Liang, Andy Zeng, Xinyun Chen, Karol Hausman, Dorsa Sadigh,
  Sergey Levine, Li~Fei-Fei, Fei Xia, and Brian Ichter.
\newblock Chain of code: Reasoning with a language model-augmented code
  emulator.
\newblock In {\em International Conference on Machine Learning}, pages
  28259--28277. PMLR, 2024.

\bibitem{li2024survey}
Yongqi Li, Xinyu Lin, Wenjie Wang, Fuli Feng, Liang Pang, Wenjie Li, Liqiang
  Nie, Xiangnan He, and Tat-Seng Chua.
\newblock A survey of generative search and recommendation in the era of large
  language models.
\newblock {\em arXiv preprint arXiv:2404.16924}, 2024.

\bibitem{luo2025rising}
Zeren Luo, Zifan Peng, Yule Liu, Zhen Sun, Mingchen Li, Jingyi Zheng, and
  Xinlei He.
\newblock The rising threat to emerging ai-powered search engines.
\newblock {\em arXiv preprint arXiv:2502.04951}, 2025.

\bibitem{lv2024codeact}
Weijie Lv, Xuan Xia, and Sheng-Jun Huang.
\newblock Codeact: Code adaptive compute-efficient tuning framework for code
  llms.
\newblock {\em arXiv e-prints}, pages arXiv--2408, 2024.

\bibitem{linkup}
Philippe Mizrahi.
\newblock
  https://www.linkup.so/blog/linkup-establishes-sota-performance-on-simpleqa,
  2025.

\bibitem{neague2025semantica}
Petru Neague, Quinten Stokkink, Naman Goel, and Johan Pouwelse.
\newblock Semantica: Decentralized search using a llm-guided semantic tree
  overlay.
\newblock {\em arXiv preprint arXiv:2502.10151}, 2025.

\bibitem{gptsearch}
OpenAI.
\newblock https://platform.openai.com/docs/models/gpt-4o-search-preview, 2025.

\bibitem{simple_evals}
OpenAI.
\newblock https://github.com/openai/simple-evals, 2025.

\bibitem{openperplex}
OpenPerplex.
\newblock https://openperplex.com, 2024.

\bibitem{perplexica}
Perplexica.
\newblock https://github.com/ItzCrazyKns/Perplexica, 2024.

\bibitem{perplexity2024}
{Perplexity AI, Inc.}
\newblock Perplexity {AI}, 2024.
\newblock https://www.perplexity.ai.

\bibitem{perplexity_sonar_pro}
PerplexityAI.
\newblock https://www.perplexity.ai/hub/blog/introducing-the-sonar-pro-api,
  2025.

\bibitem{perplexity_deep_research}
PerplexityAI.
\newblock
  https://www.perplexity.ai/hub/blog/introducing-perplexity-deep-research,
  2025.

\bibitem{robertson1995okapi}
Stephen~E Robertson, Steve Walker, Susan Jones, Micheline~M Hancock-Beaulieu,
  Mike Gatford, et~al.
\newblock Okapi at trec-3.
\newblock {\em Nist Special Publication Sp}, 109:109, 1995.

\bibitem{smolagents}
Aymeric Roucher, Albert~Villanova del Moral, Thomas Wolf, Leandro von Werra,
  and Erik Kaunismäki.
\newblock `smolagents`: a smol library to build great agentic systems.
\newblock \url{https://github.com/huggingface/smolagents}, 2025.

\bibitem{venkit2024search}
Pranav~Narayanan Venkit, Philippe Laban, Yilun Zhou, Yixin Mao, and Chien-Sheng
  Wu.
\newblock Search engines in an ai era: The false promise of factual and
  verifiable source-cited responses.
\newblock {\em arXiv preprint arXiv:2410.22349}, 2024.

\bibitem{vu2023freshllmsrefreshinglargelanguage}
Tu~Vu, Mohit Iyyer, Xuezhi Wang, Noah Constant, Jerry Wei, Jason Wei, Chris
  Tar, Yun-Hsuan Sung, Denny Zhou, Quoc Le, and Thang Luong.
\newblock Freshllms: Refreshing large language models with search engine
  augmentation, 2023.

\bibitem{wang2024survey}
Lei Wang, Chen Ma, Xueyang Feng, Zeyu Zhang, Hao Yang, Jingsen Zhang, Zhiyuan
  Chen, Jiakai Tang, Xu~Chen, Yankai Lin, et~al.
\newblock A survey on large language model based autonomous agents.
\newblock {\em Frontiers of Computer Science}, 18(6):186345, 2024.

\bibitem{wang2022self}
Xuezhi Wang, Jason Wei, Dale Schuurmans, Quoc Le, Ed~Chi, Sharan Narang,
  Aakanksha Chowdhery, and Denny Zhou.
\newblock Self-consistency improves chain of thought reasoning in language
  models.
\newblock {\em arXiv preprint arXiv:2203.11171}, 2022.

\bibitem{wei2024measuring}
Jason Wei, Nguyen Karina, Hyung~Won Chung, Yunxin~Joy Jiao, Spencer Papay,
  Amelia Glaese, John Schulman, and William Fedus.
\newblock Measuring short-form factuality in large language models.
\newblock {\em arXiv preprint arXiv:2411.04368}, 2024.

\bibitem{wei2022chain}
Jason Wei, Xuezhi Wang, Dale Schuurmans, Maarten Bosma, Fei Xia, Ed~Chi, Quoc~V
  Le, and Denny Zhou.
\newblock Chain-of-thought prompting elicits reasoning in large language
  models.
\newblock {\em Advances in Neural Information Processing Systems},
  35:24824--24837, 2022.

\bibitem{grok3}
xAI.
\newblock https://x.ai/blog/grok-3, 2025.

\bibitem{yao2022react}
Shinnosuke Yao, Joshua Zhao, Dian Yu, Nan Du, Izhak Shafran, Karthik
  Narasimhan, and Yunzhu Cao.
\newblock React: Synergizing reasoning and acting in language models.
\newblock {\em arXiv preprint arXiv:2210.03629}, 2022.

\bibitem{zheng2024gpt}
Boyuan Zheng, Boyu Gou, Jihyung Kil, Huan Sun, and Yu~Su.
\newblock Gpt-4v (ision) is a generalist web agent, if grounded.
\newblock {\em arXiv preprint arXiv:2401.01614}, 2024.

\bibitem{zuboff2019age}
Shoshana Zuboff.
\newblock The age of surveillance capitalism: The fight for a human future at
  the new frontier of power, edn.
\newblock {\em PublicAffairs, New York}, 2019.

\end{thebibliography}

\newpage
\appendix
\section*{Appendix}

\section{Example responses and comparisons from SimpleQA and FRAMES}
\label{app:examples}

\subsection{FRAMES} 

In this FRAMES example, ODS+DeepSeek-R1 correctly identifies 112 inches as the correct answer and makes an additional check with the Wolfram-Alpha API to convert the answer to 2,845mm. On the other hand, Perplexity Sonar Reasoning Pro is confused between 2,858mm and 2,845mm.

\begin{figure}[h]
\begin{lstlisting}[basicstyle=\ttfamily\tiny, frame=single]
A question from FRAMES: 
    The Jensen Interceptor (1950), produced from 1950 to 1957, had a wheelbase measurement of what in millimeters?
Correct Answer: 
    2,845 mm

Perplexity Sonar Reasoning Pro response:
    <think> Okay, let's tackle this query about the Jensen Interceptor (1950) wheelbase in millimeters. The user is asking specifically for the wheelbase measurement of the model produced between 1950 and 1957. First, I'll look through the provided search results to find mentions of the wheelbase. Let's go through each source: - **Result [1]** from automobile-catalog.com states: "Wheelbase:2858 mm / 112.5 in". That seems directly relevant. - **Result [3]** from motor-car.net lists "Wheelbase: 112 in (2,845 mm)". Hmm, slight discrepancy here-2858 vs 2845 mm. Need to check which is correct. - 
    **Result [7]** (Wikipedia) mentions a different Interceptor model (1966-1976) with a wheelbase of 107 inches (2,718 mm), but that's for a later model, so not relevant here. - **Result [8]** also from automobile-catalog.com for the 4-Litre Saloon in 1950 says "Wheelbase: 2858 mm / 112.5 in", matching Result [1]. - **Result [9]** and others don't provide wheelbase info for the 1950-1957 model. The conflicting numbers are between Results [1]/[8] (2858 mm) and Result [3] (2845 mm). Since Results [1] and [8] are both from Automobile Catalog and specify the same measurement, and Result [3] is from Motor Car History with a slightly different number, I need to assess reliability. Automobile Catalog's detailed specs in Result [1] seem more precise, including both metric and imperial conversions. The discrepancy might be due to rounding or different measurement standards, but the majority of sources point to 2858 mm. Therefore, the most accurate answer based on provided sources is 2858 mm. </think> The Jensen Interceptor produced from 1950 to 1957 had a **wheelbase of 2,858 mm** (112.5 inches)[1][8]. This measurement is consistent across technical specifications from multiple sources detailing the first-generation Interceptor's dimensions[1][3][8]. For context: - The car's overall length was 4,775 mm (188 inches)[1], making it a sizable grand tourer for its era. - The wheelbase contributed to its stable ride characteristics and spacious interior[2]. A minor discrepancy exists in one source listing the wheelbase as 2,845 mm[3], but the majority of technical documentation confirms 2,858 mm as the standard measurement[1][8].
    

    ODS-v1+DeepSeek-R1 response:
    `Initial_Thought': '<think>\nOkay, let\'s tackle this question about the Jensen Interceptor (1950) wheelbase in millimeters. The user is asking specifically about the model produced from 1950 to 1957. First, I need to go through the provided context and find any mentions of the wheelbase measurement for that specific model and years.\n\nLooking at the context snippets, there are several mentions of wheelbase measurements. Let me list out the relevant ones:\n\n1. ``Jensen Interceptor (1950) ; 4-speed manual * 112 in (2,845 mm) * 183 in (4,648 mm)." Here, 112 inches is converted to 2,845 mm. This seems to directly mention the 1950 model\'s wheelbase.\n\n2. Another snippet says: "Wheelbase: 112 in (2,845 mm). Length, 183 in (4,648 mm). Width, 66 in (1,676 mm)." Again, this matches the 1950 model\'s specifications.\n\n3. There\'s a mention of "characteristic dimensions: outside length: 4775 mm / 188 in, width: 1676 mm / 66 in, wheelbase: 2858 mm / 112.5 in; reference ...". Wait, here the wheelbase is 2858 mm, which is 112.5 inches. But the previous entries say 112 inches (2,845 mm). This discrepancy needs to be resolved.\n\nAnother context line: "Wheelbase: 2858 mm / 112.5 in. Claimed EPA passenger volume: Calculated EPA ... 1950 Jensen     Interceptor 2-Door Cabriolet Engine Horsepower/ Torque Curve ...". So here, the 1950 model\'s wheelbase is 2858 mm (112.5 inches). But earlier, it was 2,845 mm (112 inches). Which one is     correct?\n\nLooking further, another snippet says: "Technical specifications for the 1950-1957 Jensen Interceptor 2 door cabriolet ... Wheelbase: 112.2 in. Front Track: 66 in. Rear Track: 66 in     ... Nov 21, 2023". Here, the wheelbase is 112.2 inches. Converting that to mm: 112.2 inches * 25.4     mm/inch = 2849.88 mm, approximately 2850 mm. But another entry says 2858 mm (112.5 inches). There\'s inconsistency here.\n\nWait, another snippet: ``Engine, 4.0 L Austin D-Series I6 * Transmission, 4-speed manual. Dimensions. Wheelbase, 112 in (2,845 mm). Length, 183 in (4,648 mm). Width, 66 in (1,676 mm)." This clearly states 112 inches (2,845 mm). But another part says 112.5 inches (2858 mm). Which is correct?\n\nI need to check if the 1950-1957 model had different wheelbases in different years. The user is asking about the model produced from 1950 to 1957. If there\'s a variation within those years, but the context mentions both 112 and 112.5 inches. However, the majority of the entries for the 1950 model specify 112 inches (2,845 mm). The 112.5 inches (2858 mm) might be a different model year or a different variant, perhaps a later version. But the user is asking about the entire production run from 1950-1957. \n\nWait, another snippet: "characteristic dimensions: outside length: 4775 mm / 188 in, width: 1676 mm / 66 in, wheelbase:     2858 mm / 112.5 in; reference ...". This is from a section that also mentions the 4) Horsepower/Torque Curve. But the model years here aren\'t specified. However, the same context also mentions "1950 Jensen Interceptor 2-Door Cabriolet Engine Horsepower/ Torque Curve ..." which might link the 2858 mm to the 1950 model. \n\nThis is conflicting. Some sources say 2,845 mm (112 in), others 2,858 mm (112.5 in). How to resolve this? Maybe check which measurement is more commonly cited in the provided context. \n\nLooking through the context, the first mention is 112 in (2,845 mm). Then another mentions 112.2 in (approx 2850 mm). Another mentions 112.5 in (2858 mm). The technical specifications for the 1950-1957 cabriolet mention Wheelbase: 112.2 in. But another part     says 112.5 in. \n\nAlternatively, perhaps there\'s confusion between different models. The Jensen Interceptor name was used again in the 1960s, but the user is asking about'}, {'Thought': "The context mentions both 112 inches (2844.8 mm) and 112.5 inches (2857.5 mm). The 1950 model's original specs from the article state 112 inches (2845 mm), while later references to the cabriolet and other variants might use 112.5 inches. Since the user specified the 1950-1957 model, the primary value from the 1950 article is 112 inches. Let me convert this value."}, {'Action':     'calculate'}, {'Action_Input': '112 inches to millimeters'}, {'Observation': {'2845mm'}},     {'Thought': ''}, {'Action': 'Done'},'final_result': 'The wheelbase measurement of the Jensen Interceptor (1950) produced from 1950 to 1957 is 2,845 millimeters.'' 
    \end{lstlisting}
    \end{figure} 

In the below Frames example, \osa+Llama3.1-70B correctly identifies Dune Part Two as the film in question, while Perplexity Sonar Reasoning Pro \cite{perplexity_sonar_pro} and \osa+DeepSeek-R1 is unable to determine the specific answer.

\begin{figure}[h!]
    \centering
    \begin{lstlisting}[basicstyle=\ttfamily\tiny, frame=single]
A question from FRAMES: 
    Which film featuring a solar eclipse in its opening scene is adapted from the same source material as a David Lynch movie?
Correct Answer: 
    Dune Part Two
Perplexity Sonar Reasoning Pro:
    <think> Okay, let's tackle this query. The user is asking which film featuring a solar eclipse in its opening scene is adapted from the same source material as a David Lynch movie. First, I need to identify David Lynch movies that are adaptations. From the search results, I see that David Lynch directed the 1984 adaptation of *Dune* [1][3][4][7][9]. The source material for that is Frank Herbert's novel *Dune*. Now, looking for another film adapted from the same novel (*Dune*) that has a solar eclipse in its opening scene. The search results mention Denis Villeneuve's 2021 *Dune: Part One* and the upcoming *Dune: Part Two*. Let me check the trailers and scenes described. In search result [1], the trailer breakdown for *Dune: Part Two* mentions a scene with blood orange hues, possibly due to an environmental effect like an eclipse. However, it's not explicitly stated as the opening scene. Search result [3] talks about Villeneuve's 2021 *Dune* but doesn't mention an eclipse in the opening. Wait, search result [7] is about how David Lynch's *Dune* redefined the saga. Maybe there's a connection here. But the user is asking for another film with a solar eclipse opening, same source as Lynch's. Looking at search result [9], the new *Dune* trailer uses Pink Floyd's "Eclipse," which might be a nod to an eclipse scene. The 2021 *Dune* film by Villeneuve is part one, and part two might have such a scene. However, the user specified "opening scene." Wait, search result [1] describes the trailer opening with Paul and Chani on Arrakis....Given the information, the most accurate answer would be that the film in question is likely one of the adaptations of "Dune," but the specific detail about the solar eclipse in the opening scene is not clearly confirmed in the provided sources.

ODS-v1+Llama3.1-70B response: 
    '$Initial\_Thought$'
        To answer this question, I\'ll follow these steps:
        1. Identify the source material that David Lynch adapted into a film.
        2. Find a film featuring a solar eclipse in its opening scene that is adapted from the same source material.
        From the given context, I can see that David Lynch adapted the novel "Dune" into a film in 1984.
        Next, I\'ll look for a film featuring a solar eclipse in its opening scene that is also adapted from the novel "Dune", 
    {'Thought': ''}, 
    {'Action': 'search\_internet'}, 
    {'Action\_Input': 'film featuring solar eclipse in opening scene adapted from Dune'},
    {'Observation': Dune: Part Two opening scene features an ambush between Harkonnen soldiers and Fremen against the backdrop of a visually striking solar eclipse ... Apr 9, 2024', "Paul 'eclipses' the Golden Path by being the wrong son born to the wrong mother at the wrong time, and thus, all is in tune, but the sun is eclipsed by the ... Sep 10, 2020"},
    {'Thought': 'Based on the search results, I found that the film featuring a solar eclipse in its opening scene that is adapted from the same source material as a David Lynch movie is "Dune: Part Two" (2024). The source material is the novel "Dune" by Frank Herbert, which was also adapted into a film by David Lynch in 1984. Do I need further steps? No.'}, 
    {'Action': 'Done'}'

ODS-v1+DeepSeeek-R1 response:
    'final\_result': 'The film "Dune" (1984), directed by David Lynch and adapted from Frank Herbert\'s novel, shares its source material with Denis Villeneuve\'s "Dune" adaptations. However, the film featuring a solar eclipse in its opening scene from the same context is "2001: A Space Odyssey" (1968), though it is not related to Lynch\'s work. Based on the provided context, the answer reconciling both criteria is **"Dune" (1984)**, as it is Lynch\'s adaptation of Herbert\'s novel, though no solar eclipse is mentioned in its opening.', 'react\_sequence': [{'Initial\_Thought': '<think>\nOkay, let\'s tackle this question step by step. The user is asking which film that features a solar eclipse in its opening scene is adapted from the same source material as a David Lynch movie. \n\nFirst, I need to recall the context provided. The context mentions several David Lynch films, including his 1984 adaptation of "Dune." It also talks about other films with solar eclipses, like "2001: A Space Odyssey" which opens with an eclipse. The key here is to find a connection between a solar eclipse in a film\'s opening and a David Lynch adaptation.\n\nDavid Lynch directed "Dune" (1984), which is based on Frank Herbert\'s novel. The context also states that Denis Villeneuve\'s "Dune" is another adaptation, split into two parts. However, the user is looking for a film with a solar eclipse in the opening that shares the same source material as Lynch\'s movie. \n\nLooking at the solar eclipse films mentioned, "2001: A Space Odyssey" (1968) opens with an eclipse. But that\'s based on Arthur C. Clarke\'s work, not related to Lynch\'s "Dune." Another mention is "Barabbas" (1961) which used a real eclipse during filming, but that\'s a biblical story. \n\nWait, the context also mentions "Dune" adaptations. Lynch\'s "Dune" is based on Herbert\'s book. The other adaptation by Villeneuve is also from the same source. But does Villeneuve\'s "Dune" have a solar eclipse in the opening? The context doesn\'t specify that. However, "2001: A Space Odyssey" is noted for its eclipse opening but isn\'t related to "Dune." \n\nWait, maybe the user is conflating two different elements: a film with a solar eclipse opening and a Lynch adaptation. Since Lynch\'s "Dune" is the key here, and the other adaptation is Villeneuve\'s, but neither of them have a solar eclipse in their openings. The context does mention "2001: A Space Odyssey" as having an eclipse opening, but it\'s not related to Lynch\'s source material. \n\nWait, perhaps the answer is that there isn\'t a direct connection, but the user might be thinking of "Dune" adaptations. However, the context mentions that Villeneuve\'s "Dune" is split into two parts, avoiding Lynch\'s issue. But the solar eclipse isn\'t in "Dune." \n\nWait, another angle: the context mentions "Bloody Birthday" (1981) where a lunar eclipse is part of the plot, but that\'s a horror film. The 1907 film "The Eclipse" is about an eclipse but unrelated. \n\nWait, the user\'s question is tricky. They want a film that both has a solar eclipse in the opening and is adapted from the same source as a Lynch movie. Lynch\'s "Dune" is from Frank Herbert\'s book. Are there other adaptations of "Dune" that have a solar eclipse opening? The 2000 miniseries by John Harrison? The context doesn\'t mention an eclipse there. \n\nAlternatively, maybe the answer is "2001: A Space Odyssey" but that\'s not based on the same source as Lynch\'s movie. Wait, no. The only common source material for Lynch\'s movie is "Dune." So unless another "Dune" adaptation has a solar eclipse opening, which isn\'t mentioned in the context, perhaps the answer is that there isn\'t one. But the context does mention "2001: A Space Odyssey" as having an eclipse opening, and it\'s directed by Kubrick. But that\'s unrelated to Lynch\'s work....}, 
    {'Action': 'Done'}, {'Action\_Input': ''}
    \end{lstlisting}
\end{figure}

In this example from FRAMES, \osa+Llama3.1-70B searches again once it finds that the company which launched the commercial was Levi \& Strauss. This leads the agent to correctly identify Dockers as the first branch launched by Levi \& Strauss after they changed the name of Meat Loaf in 1984. Meanwhile, Perplexity \cite{perplexity2024} is unable to determine the answer from a single search. 
\begin{figure}[h!]
        \centering
        \begin{lstlisting}[basicstyle=\ttfamily\tiny, frame=single]
A question from FRAMES: 
    After Meat Loaf legally changed his name due to a commercial, what was the first new brand launched by the company to whom the commercial belonged?
Correct Answer:
    Dockers
Perplexity response:
    The information provided does not link the name change of Meat Loaf (from Marvin Lee Aday to Michael Lee Aday) to a specific commercial or a company that launched a new brand subsequent to this name change. Meat Loaf changed his first name from Marvin to Michael, but the reason given was due to a jeans commercial, not because of any direct involvement with a company that then launched a new brand[5]. There is no additional information in the sources provided that connects this name change to the launch of a new brand by any company. 
ODS-v1+Llama3.1-70B response:
    '$Initial\_Thought$':
        To answer this question, I'll follow these steps:
        1. Identify the company that owned the commercial that led to Meat Loaf's name change.
        2. Research the company's product launches after the name change.
      From the context, I found that the commercial was from Levi Strauss & Co. According to the text, Meat Loaf changed his name from Marvin to Michael in 1984 due to a Levi's commercial.
      Next, I'll search for information on new brands launched by Levi Strauss & Co. after 1984.",
   'Action': 'search\_internet', 'Action\_Input': 'Levi Strauss & Co. new brands launched after 1984',
    'Observation': 
        'The Dockers brand launches sunglasses, bed linens, & bath items. The Dockers brand introduces innovations such as the Never Iron & Thermal Adapt. 
    'Thought': 
        "Based on the search results, I found that Levi Strauss & Co. launched the Dockers brand in 1986, which was the first new brand launched by the company after Meat Loaf's name change in 1984. Do I need further steps? No.", 
    'Action': 'Done'
\end{lstlisting}
\end{figure}

In this example from FRAMES, \osa+Llama3.1-70B uses the Wolfram calculator tool to report the correct result of 2 years. On the other hand, Perplexity reports 2 years and 4 months in the wrong format since the question explicitly asked for the result in years.
\begin{figure}[h!]
    \centering
    \begin{lstlisting}[basicstyle=\ttfamily\tiny, frame=single]
A question from FRAMES: 
    How many years elapsed between the release of the song "I Think I'm Go Go" by the band Squeeze and the theatrical premier of E.T. the movie?
Correct Answer: 
    2 years
Perplexity response:
    The song "I Think I'm Go Go" by Squeeze was released on February 4, 1980[1][4]. The theatrical premiere of the movie "E.T. the Extra-Terrestrial" was on June 10, 1982, in Los Angeles, California[2]. Therefore, the elapsed time between the release of the song and the movie premiere is approximately 2 years and 4 months.
ODS-v1+Llama3.1-70B response: 
    '$Initial\_Thought$'''To answer this question, I\'ll need to follow these steps:\n\n1. Identify the release year of the song "I Think I\'m Go Go" by Squeeze.\n2. Identify the theatrical premiere year of the movie E.T.\n3. Calculate the difference between the two years to find out how many years elapsed between the two events.\n\nLet\'s start by looking for the release year of the song "I Think I\'m Go Go" by Squeeze..., {'Thought': 'According to the search results, the song "I Think I\'m Go Go" by Squeeze was released in 1980. I need to find the theatrical premiere year of the movie E.T. I will calculate the difference between the two years to find out how many years elapsed between the two events. Do I need further steps? No.'},  Do I need further steps? No.', {'Action': calculate}, {'Action_Input': '1982-1980'}'Thought': 'final\_result': 'Cid Corman was born in 1924, so as of the publication date of Rupi Kaur\'s book "Milk and Honey" (November 4, 2014), he would have been 90 years old. Do I need further steps? No, {'Action': 'Done'}{'final_result': 'The song "I Think I\'m Go Go" by Squeeze was released in 1980, and the theatrical premiere of the movie E.T. was in 1982. Therefore, 2 years elapsed between the release of the song and the theatrical premiere of the movie.'

    \end{lstlisting}
\end{figure}

In this example from FRAMES, \osa+DeepSeek-R1 is encouraged to continue thinking and utilizes chain-of-thought to produce the correct answer by the Open Reasoning Agent. By reflecting on its own answer and feeding back its reflection to the language model, \osasp is able to find the correct answer that 31 years passed since Starbucks introduced juice-blended Frappuccinos and when the company that is accredited for inventing the Frappuccinos is founded. Perplexity Sonar Reasoning Pro \cite{perplexity_sonar_pro} fails to answer this question and concludes that there wasn’t enough information provided in the context.

\begin{figure}[h!]
    \centering
    \begin{lstlisting}[basicstyle=\ttfamily\tiny, frame=single]
A question from FRAMES: 
    How many years before Starbucks introduced juice-blended Frappuccinos, was the company that is accredited for inventing the Frappuccino, founded?
Correct Answer: 
     31 years 
Perplexity Sonar Reasoning Pro response:
    <think> Okay, let's tackle this query step by step. The user is asking how many years before Starbucks introduced juice-blended Frappuccinos, the company credited with inventing the Frappuccino was founded. First, I need to identify the company that invented the Frappuccino. From the search results, it's clear that George Howell's The Coffee Connection created the original Frappuccino....So assuming CC existed since at least late '80s or early '90s but without exact date. Given this ambiguity and lack of explicit founding date I cannot answer the question.
ODS-v1+DeepSeek-R1 response: 
    'Initial_Thought': '<think>\nOkay, let\'s tackle this question step by step. The user is asking how many years before Starbucks introduced juice-blended Frappuccinos the company credited with inventing the Frappuccino was founded.\n\nFirst, I need to identify the key pieces of information required. The two main dates here are: 1) the year Starbucks introduced juice-blended Frappuccinos, and 2) the founding year of the company that originally invented the Frappuccino....
    {'Thought': "I need to find the year Starbucks introduced juice-blended Frappuccinos and the founding year of The Coffee Connection (the original inventor of Frappuccino). Now, I need to determine when The Coffee Connection (credited with inventing Frappuccino) was founded. Calculate the time difference between 1975 (The Coffee Connection's founding) and 2006 (juice-blended Frappuccino introduction). The answer is derived using external knowledge for The Coffee Connection's founding year, as the context lacks this detail."}, {'Action': 'Done'}, {'Action_Input': ''}'final_result': 'The company credited with inventing the Frappuccino, The Coffee Connection, was founded in 1975. Starbucks introduced juice-blended Frappuccinos in 2006, which is 31 years later. Therefore, the difference is 31 years
    \end{lstlisting}
\end{figure}

In the next example  question and answer from FRAMES, the \osa+DeepSeek-R1 is encouraged to continue thinking via the ReAct Thought/Action/Observation sequence and utilizes chain-of-thought to produce the correct answer. The name of the Japanese man who protested the U.S.'s involvement in the Iraq War, and who has also been awarded multiple times the same award that "Shrek" won in 2002, beating "Jimmy Neutron: Boy Genius", is Miyazaki Hayao. On the other hand, Perplexity Sonar Reasoning Pro fails to answer this question and concludes that there wasn’t enough information provided in the context.

\begin{figure}[h!]
    \centering
    \begin{lstlisting}[basicstyle=\ttfamily\tiny, frame=single]
A question from FRAMES: 
    What is the name of the Japanese man who protested the U.S.'s involvement in the Iraq War, and who has also been awarded multiple times the same award that "Shrek" won in 2002, beating "Jimmy Neutron: Boy Genius"?
Correct Answer: 
     Hayao Miyazaki 
Perplexity Sonar Reasoning Pro response:
     <think> Okay, let's tackle this query. The user is asking for the name of a Japanese man who protested the U.S. involvement in the Iraq War and also won the same award that "Shrek" won in 2002, beating "Jimmy Neutron: Boy Genius." First, I need to figure out what award "Shrek" won in 2002. A quick recall tells me that the first Shrek movie was released in 2001, and it won the first Academy Award for Best Animated Feature in 2002. The other nominee      that year was "Jimmy Neutron: Boy Genius." So the award in question is the Oscar for Best Animated Feature. Next, I need to find a Japanese man involved in protests against the Iraq War who has also won this Oscar. Looking through the provided search results, most of them discuss Japanese anti-war protests, particularly focusing on individuals like Obora Toshiyuki and others from the Tachikawa Tent Village group. However, none of these sources mention any connection to film awards. The user might be conflating two different people or there's a mix-up in the details.      The search results don't indicate any link between the anti-war protesters and the Academy Awards. It's possible that the user has confused a protester with someone else in the film industry. Alternatively, there could be a misunderstanding about the awards won by individuals involved in protests. Given that none of the search results mention an Oscar winner among the protesters, it's likely that there's no such person as described. The protesters mentioned, like Obora Toshiyuki, were recognized for their activism but not for film achievements. Therefore, the answer should clarify that there's no known Japanese protester against the Iraq War who also won the mentioned      Oscar. </think> The query appears to contain contradictions or factual errors based on      available evidence. Here's why: **Regarding Anti-Iraq War Protests in Japan:** -      Japanese antiwar activists like **Obora Toshiyuki** and his colleagues from the      **Tachikawa Tent Village** group were arrested in 2004 for distributing antiwar flyers      opposing Japan's Self-Defense Forces deployment to Iraq [1][5][8]. Amnesty International designated them as Japan's first      "Prisoners of Conscience"[3]. **Regarding Awards:** - *Shrek* (2001) won the **first Academy Award for Best Animated Feature** in 2002, defeating *Jimmy Neutron: Boy Genius*. No Japanese individual or film connected to Iraq      War protests has won this award. **Key Discrepancies:** 1. No evidence links Obora Toshiyuki or other Japanese antiwar activists to animation awards. 2. The Academy Award category referenced (Best Animated Feature) did not      exist before 2001, and Japanese films/artists have rarely won it (*Spirited Away* won in 2003 but is unrelated to Iraq protests). This suggests either a conflation of unrelated facts or a fictional premise. For further clarification, additional context about the alleged award connection would be needed.  
ODS-v1+DeepSeek-R1 response: 
     'Initial_Thought': '<think>\nOkay, let\'s try to figure this out. The user is asking for the name of a Japanese man who protested the U.S.\'s involvement in the Iraq War and also won the same award that "Shrek" won in 2002, which beat "Jimmy Neutron: Boy Genius." \n\nFirst, I need to recall that the award in question here is the Academy Award for Best Animated Feature. "Shrek" won the first Oscar in that category in 2002, and "Jimmy Neutron: Boy Genius" was one of the other nominees that year. So the award is the Oscar for Best Animated Feature.\n\nNext, the person we\'re looking for is Japanese, involved in protests against the Iraq War, and has won this award multiple times. From the context provided, there are mentions of Japanese antiwar protesters, specifically three activists who were arrested for distributing antiwar flyers. However, those individuals are referred to as "Prisoners of Conscience" by Amnesty International, but their names aren\'t given in the context snippets.\n\nLooking at the Oscar-related context, there\'s mention of Studio Ghibli and Hayao Miyazaki. For example, "Spirited Away" won the Oscar for Best Animated Feature in 2003, which is after the category was introduced in 2001. Miyazaki\'s films have won multiple times, including "Spirited Away" and "The Boy and the Heron" in 2024. The context also states that Studio Ghibli has the most wins (two) and nominations (seven). \n\nNow, connecting the dots: the user is asking for someone who both protested the Iraq War and won the Best Animated Feature Oscar multiple times. Hayao Miyazaki is a prominent Japanese filmmaker known for his anti-war stance. The context mentions that Miyazaki didn\'t attend the 2003 Oscars because of his opposition to the Iraq War. Additionally, his films have won the Best Animated Feature award twice ("Spirited Away" and "The Boy and the Heron"), which fits the "multiple times" part. \n\nHowever, the context also mentions that "Shrek" was the first winner in 2002, and Miyazaki\'s first win was in 2003. The user\'s question specifies that the person has won the same award that "Shrek" won, which Miyazaki did, albeit in subsequent years. The protest connection is there because Miyazaki opposed the Iraq War, as indicated in the context where he didn\'t attend the Oscars due to this opposition. \n\nTherefore, putting it all together, the answer should be Hayao Miyazaki. He is the Japanese filmmaker known for his anti-war views (including opposition to the Iraq War) and has won the Academy Award for Best Animated Feature multiple times, which is the same award "Shrek" won in 2002.\n</think>\n\nThe Japanese individual in question is **Hayao Miyazaki**. \n\n**Step-by-Step Explanation:** \n1. **Award Context:** The Academy Award for Best Animated Feature was first awarded in 2002 (*Shrek* won, beating *Jimmy Neutron: Boy Genius*). \n2. **Japanese Oscar Winners:** Studio Ghibli co-founder Hayao Miyazaki won this award twice: for *Spirited Away* (2003) and *The Boy and the Heron* (2024). \n3. **Anti-War Stance:** Miyazaki publicly opposed the 2003 Iraq War. The context notes he boycotted the 2003 Oscars (where *Spirited Away* won) due to his anti-war principles. \n\nThus, Miyazaki matches both criteria: protesting the Iraq War and winning the same Oscar category as *Shrek*.'}, {'Thought': "I need to identify the Japanese man who protested the U.S. involvement in the Iraq War and also won the Best Animated Feature Oscar multiple times. Verify Miyazaki's Oscar wins. Confirm his anti-war stance. I now know the final answer."}, {'Action': 'Done'}, {'Action_Input': ''} 'final_result': 'Hayao Miyazaki',   

    \end{lstlisting}
\end{figure}

\newpage
\subsection{SimpleQA}

\osa+DeepSeek-R1 correctly follows a reasoning path which distinguishes July 21, 2022 as the correct date that Kaitlin Armstrong pleaded not guilty to the murder charge of Moriah Wilson and was arraigned. The \osasp agent cross checks the two conflicting dates of July 20 and July 21, 2022 and chooses July 21 correctly. Conversely, Perplexity Pro gets confused and gives the the wrong answer of July 20, 2022, lacking the additional check which allowed /osasp to answer the query. 

\begin{figure}[h!]
    \centering
    \begin{lstlisting}[basicstyle=\ttfamily\tiny, frame=single]
A question from SimpleQA: 
    What is the month, day, and year Kaitlin Armstrong pleaded not guilty to the murder charge of Moriah Wilson and was arraigned?}
Correect answer:    
    July 21, 2022 

Perplexity Sonar Reasoning Pro response:
<think> Okay, so I need to find the exact date when Kaitlin Armstrong pleaded not guilty to the murder charge of Moriah Wilson and was arraigned. Let me go through the search results provided to pinpoint this information. Starting with the first search result [1], it mentions that on July 20, 2022, Armstrong had her first court appearance where she pleaded not guilty. The source is CBS Austin, which seems reliable. Another mention in [4] from CBS Austin's timeline confirms that on July 20, 2022, Armstrong entered a not guilty plea and requested a speedy trial.  Looking at [6], ESPN's article also states that after being apprehended in Costa Rica and returned to the U.S., Armstrong pleaded not guilty. The timeline in [4] lists July 20 under 2022 events, specifying her first court appearance where she pleaded not guilty.  Additionally, [7] from CBS News mentions that Armstrong had pleaded not guilty to first-degree murder, but doesn't specify the date. However, combining all these sources, the consistent date across multiple references is July 20, 2022.  I should cross-check if any other dates are mentioned elsewhere. For example, [2] talks about the trial timeline starting in October 2023 but doesn't mention the arraignment date. [8] and [9] discuss later events like appeals and escape attempts but not the initial plea.  Therefore, consolidating the information from [1], [4], and [6], the correct date is July 20, 2022. </think>
Kaitlin Armstrong pleaded not guilty to the murder charge of Moriah Wilson and was arraigned on **July 20, 2022**[1][4][6]. This occurred during her first court appearance after being extradited to the U.S., where she also requested a speedy trial[4]. The Travis County court processed her plea following her arrest in Costa Rica and subsequent return to Texas custody[6].

ODS-v1+DeepSeek-R1 response: 
    {'Initial_Thought': '<think>\nOkay, let\'s tackle this query step by step. The user is asking for the specific date (month, day, and year) when Kaitlin Armstrong pleaded not guilty to the murder charge of Moriah Wilson and was arraigned.\n\nFirst, I need to recall the context provided. The context includes multiple dates and events related to Kaitlin Armstrong\'s case. Let me go through each relevant piece of information.\n\nLooking at the context entries, several mention her plea. For example, one entry says: "In July 2022, Armstrong pleaded not guilty to Wilson\'s murder. Her trial date was initially scheduled to begin on June 26, 2023. However, on ... Aug 27, 2024". Another entry states: "Armstrong was arraigned on July 21, 2022, pleading not guilty to the murder charge...., 
    {'Thought': 'I need to resolve the conflicting dates. The context mentions both July 20 and July 21, 2022, for her plea. Let me check more entries. Another entry states: "Kaitlin Armstrong was arraigned on July 21, 2022, pleading not guilty to the murder charge. She was held on a $3.5 million bond..." This explicitly links the arraignment to July 21.....Given the explicit mention of arraignment on July 21, 2022, in the context, that\'s likely the correct date. Therefore, the answer is July 21, 2022.'}, 
    {'Action': 'Done'}, {'Action_Input': ''}
    'final_result': 'Kaitlin Armstrong pleaded not guilty to the murder charge of Moriah Wilson and was arraigned on July 21, 2022.' 
\end{lstlisting}
\end{figure} 

\osa+DeepSeek-R1 correctly follows a reasoning path which identifies segment producer as the role Rosie Perez held in the TV show, In Living Color. \osasp cross checks multiple sources to confirm the result. Conversely, Perplexity Sonar REasoning Pro talks about her “directing segments” and fails to organise its reasoning into the correct response.

\begin{figure}[h!]
    \centering
    \begin{lstlisting}[basicstyle=\ttfamily\tiny, frame=single]
A question from SimpleQA: 
    Other than being a choreographer for the TV series In Living Color, what other job did Rosie Perez do on the show? 
Correct answer: 
    segment producer

Perplexity Sonar Reasoning Pro response: 
    <think> Okay, let's tackle this query about Rosie Perez and her roles on ``In Living Color." The user wants to know what other job she did besides being a choreographer. First, I need to go through the search results provided. Let's start by skimming through each source. From the search results, [1] mentions Rosie Perez as part of the Fly Girls collaboration. [2] states she was a cast member and choreographer. ...However, other sources like [9] don't list her as part of the main cast but as a crew member. The IMDb entry in [8] shows she directed some segments of the show. Therefore, besides being a choreographer, Rosie Perez also directed certain segments of ``In Living Color," specifically the dance bumpers. Additionally, if being a cast member refers to appearing in sketches, but that's less clear from the sources..... she **directed segments** of the show[8]. Specifically, Perez directed the ``Dance Bumpers" segments during Seasons 2-4 (1991-1993)[8]. These short dance sequences transitioned between sketches and commercials, showcasing her creative vision beyond choreography....While not a regular sketch performer herself[9], her behind-the-scenes influence extended into production decisions and cultural curation[7].

ODS-v1+DeepSeek-R1 response: 
    {Initial_Thought': '<think>\nOkay, let\'s tackle this question. The user is asking what other job Rosie Perez did on the TV show "In Living Color" besides being a choreographer. I need to go through the provided context to find any mentions of her roles beyond choreography.
    Looking at the context snippets, several mention her as the choreographer for The Fly Girls. But there\'s a line that says: "Career ; In Living Color, and also worked as a segment producer....Also, another line: "Career ; In Living Color, and also worked as a segment producer." So multiple mentions of her being a segment producer... This is explicitly mentioned in the context snippets, which highlight her dual contributions to the show.'}, 
    {'Thought': ''}, {'Action': 'Done'}, {'Action_Input': ''}
    'final_result': 'Rosie Perez also worked as a segment producer on the TV series "In Living Color".'

\end{lstlisting}
\end{figure}

\newpage
\section{Question and agentic response examples for few-shot chain-of-thought to be used in the ReAct-based Open Reasoning Agent of \osa-v1}
\label{app:prompts}

The following 20 examples we used in few-shot chain-of-thought were derived from a community campaign we ran to improve ODS with better prompts.

\paragraph{Example 1}
\begin{verbatim}
<Question> What historical event happened closest in time to the invention of the 
telephone: the American Civil War or the establishment of the Eiffel Tower? </Question>
<Thought> I need to find the year the telephone was invented. </Thought>
<Action>search_internet</Action><Action_Input>year of telephone invention</Action_Input>
<Observation> The telephone was invented in 1876. </Observation>
<Thought> I need to find the year the American Civil War ended. </Thought>
<Action>search_internet</Action><Action_Input>year American Civil War ended</Action_Input>
<Observation> The American Civil War ended in 1865. </Observation>
<Thought> I need to find the year the Eiffel Tower was established. </Thought>
<Action>search_internet</Action><Action_Input>year Eiffel Tower established</Action_Input>
<Observation> The Eiffel Tower was completed in 1889. </Observation>
<Thought> I will calculate the time differences from 1876. </Thought>
<Action>calculate</Action><Action_Input>|1876 - 1865| and |1889 - 1876|</Action_Input>
<Observation> 11 years (Civil War) and 13 years (Eiffel Tower). </Observation>
<Thought> The event closest in time to the invention of the telephone is
the end of the American Civil War. </Thought>
<Action>Done</Action>
<Final_Answer> The historical event closest in time to the invention of the 
telephone is the end of the American Civil War (11 years apart). </Final_Answer>
\end{verbatim}

\paragraph{Example 2}
\begin{verbatim}
<Question> Which country has a higher population density: Japan or India? </Question>
<Thought> I need to find the population and area of Japan. </Thought>
<Action>search_internet</Action>
<Action_Input>population and area of Japan</Action_Input>
<Observation> Japan has a population of 125 million and an area of 
377,975 square kilometers. </Observation>
<Thought> I need to find the population and area of India. </Thought> 
<Action>search_internet</Action>
<Action_Input>population and area of India</Action_Input>
<Observation> India has a population of 1.38 billion and an area of 
3,287,263 square kilometers. </Observation>
<Thought> I need to calculate the population densities of Japan and India. </Thought>
<Action>calculate</Action>
<Action_Input>125 million / 377,975 and 1.38 billion / 3,287,263</Action_Input>
<Observation> Japan: 330.7 people/km²; India: 419.6 people/km². </Observation>
<Thought> India has a higher population density than Japan. </Thought>
<Action>Done</Action>
<Final_Answer> India has a higher population density (419.6 people/km²) 
than Japan (330.7 people/km²). </Final_Answer>
\end{verbatim}

\paragraph{Example 3} 
\begin{verbatim}
<Question> Which country has won more total Olympic gold medals: 
the United States or China? </Question>
<Thought> I need to find the total Olympic gold medals won by 
the United States. </Thought>
<Action>search_internet</Action>
<Action_Input>total Olympic gold medals won by the United States</Action_Input>
<Observation> The United States has won 1,127 gold medals. </Observation>
<Thought> I need to find the total Olympic gold medals won by China. </Thought>
<Action>search_internet</Action>
<Action_Input>total Olympic gold medals won by China</Action_Input>
<Observation> China has won 283 gold medals. </Observation>
<Thought> I will compare the two numbers. </Thought>
<Action>calculate</Action>
<Action_Input>1,127 - 283</Action_Input>
<Observation> The United States has 844 more gold medals than China. </Observation>
<Thought> The United States has won more gold medals. </Thought>
<Action>Done</Action>
<Final_Answer> The United States has won more Olympic gold medals 
(1,127) than China (283). </Final_Answer>
\end{verbatim}

\paragraph{Example 4}
\begin{verbatim}
<Question> Who discovered the structure of DNA, and in which year was 
the discovery made? </Question>
<Thought> I need to find the scientists who discovered the structure 
of DNA. </Thought>
<Action>search_internet</Action>
<Action_Input>scientists who discovered DNA structure</Action_Input>
<Observation> James Watson and Francis Crick discovered the structure 
of DNA. </Observation>
<Thought> I need to find the year of their discovery. </Thought>
<Action>search_internet</Action>
<Action_Input>year DNA structure discovered</Action_Input>
<Observation> The structure of DNA was discovered in 1953. </Observation>
<Thought> Do I need further steps? No. </Thought>
<Action>Done</Action>
<Final_Answer> James Watson and Francis Crick discovered the structure 
of DNA in 1953. </Final_Answer>
\end{verbatim}

\paragraph{Example 5}
\begin{verbatim}
<Question> How many meters taller is the Burj Khalifa compared to 
the Empire State Building? </Question>
<Thought> I need to find the height of the Burj Khalifa. </Thought>
<Action>search_internet</Action>
<Action_Input>height of Burj Khalifa</Action_Input>
<Observation> The Burj Khalifa is 828 meters tall. </Observation>
<Thought> I need to find the height of the Empire State Building. </Thought>
<Action>search_internet</Action>
<Action_Input>height of Empire State Building</Action_Input>
<Observation> The Empire State Building is 381 meters tall. </Observation>
<Thought> I will calculate the difference in height. </Thought>
<Action>calculate</Action>
<Action_Input>828 - 381</Action_Input>
<Observation> The difference is 447 meters. </Observation>
<Thought> Do I need further steps? No. </Thought>
<Action>Done</Action>
<Final_Answer> The Burj Khalifa is 447 meters taller than 
the Empire State Building. </Final_Answer>
\end{verbatim}

\paragraph{Example 6}
\begin{verbatim}
<Question> Which country launched the first satellite into space, 
and what was the name of the satellite? </Question>
<Thought> I need to find the country that launched the first satellite 
into space. </Thought>
<Action>search_internet</Action>
<Action_Input>first satellite launched into space</Action_Input>
<Observation> The Soviet Union launched the first satellite. </Observation>
<Thought> I need to find the name of the satellite. </Thought>
<Action>search_internet</Action>
<Action_Input>name of first satellite in space</Action_Input>
<Observation> The first satellite was Sputnik 1. </Observation>
<Thought> Do I need further steps? No. </Thought>
<Action>Done</Action>
<Final_Answer> The Soviet Union launched the first satellite into 
space, named Sputnik 1. </Final_Answer>
\end{verbatim}

\paragraph{Example 7}
\begin{verbatim}
<Question> Which novel by George Orwell introduced the concept of 
'Big Brother,' and in what year was it published? </Question>
<Thought> I need to identify the novel by George Orwell that 
introduced 'Big Brother.' </Thought>
<Action>search_internet</Action>
<Action_Input>novel by George Orwell Big Brother</Action_Input>
<Observation> The novel is '1984.' </Observation>
<Thought> I need to find the year it was published. </Thought>
<Action>search_internet</Action>
<Action_Input>year '1984' by George Orwell published</Action_Input>
<Observation> '1984' was published in 1949. </Observation>
<Thought> Do I need further steps? No. </Thought>
<Action>Done</Action>
<Final_Answer> George Orwell's novel '1984,' which introduced 
the concept of 'Big Brother,' was published in 1949. </Final_Answer>
\end{verbatim}

\paragraph{Example 8}
\begin{verbatim}
<Question> Which country hosted the first FIFA World Cup, and in 
what year? </Question>
<Thought> I need to find the country that hosted the first FIFA 
World Cup. </Thought>
<Action>search_internet</Action>
<Action_Input>country hosted first FIFA World Cup</Action_Input>
<Observation> Uruguay hosted the first FIFA World Cup. </Observation>
<Thought> I need to find the year of the first FIFA World Cup. </Thought>
<Action>search_internet</Action>
<Action_Input>year of first FIFA World Cup</Action_Input>
<Observation> The first FIFA World Cup was held in 1930. </Observation>
<Thought> Do I need further steps? No. </Thought>
<Action>Done</Action>
<Final_Answer> Uruguay hosted the first FIFA World Cup in 1930. </Final_Answer>
\end{verbatim}

\paragraph{Example 9}
\begin{verbatim}
<Question> Who invented the light bulb, and what company did he 
later establish? </Question>
<Thought> I need to find who invented the light bulb. </Thought>
<Action>search_internet</Action>
<Action_Input>inventor of the light bulb</Action_Input>
<Observation> Thomas Edison invented the light bulb. </Observation>
<Thought> I need to find the company Thomas Edison established. </Thought>
<Action>search_internet</Action>
<Action_Input>company founded by Thomas Edison</Action_Input>
<Observation> Thomas Edison founded General Electric. </Observation>
<Thought> Do I need further steps? No. </Thought>
<Action>Done</Action>
<Final_Answer> Thomas Edison invented the light bulb and later 
established General Electric. </Final_Answer>
\end{verbatim}

\paragraph{Example 10}
\begin{verbatim}
<Question> In which city was the Declaration of Independence signed, 
and in what building? </Question>
<Thought> I need to find the city where the Declaration of Independence 
was signed. </Thought>
<Action>search_internet</Action>
<Action_Input>city where Declaration of Independence was signed</Action_Input>
<Observation> The Declaration of Independence was signed in Philadelphia. </Observation>
<Thought> I need to find the building where it was signed. </Thought>
<Action>search_internet</Action>
<Action_Input>building where Declaration of Independence was signed</Action_Input>
<Observation> It was signed in Independence Hall. </Observation>
<Thought> Do I need further steps? No. </Thought>
<Action>Done</Action>
<Final_Answer> The Declaration of Independence was signed in 
Philadelphia at Independence Hall. </Final_Answer>
\end{verbatim}

\paragraph{Example 11}
\begin{verbatim}
<Question> Who developed the theory of general relativity, 
and in what year was it published? </Question>
<Thought> I need to find who developed the theory of general 
relativity. </Thought>
<Action>search_internet</Action>
<Action_Input>developer of general relativity</Action_Input>
<Observation> Albert Einstein developed the theory of general 
relativity. </Observation>
<Thought> I need to find the year it was published. </Thought>
<Action>search_internet</Action>
<Action_Input>year general relativity published</Action_Input>
<Observation> The theory of general relativity was published in 1915. </Observation>
<Thought> Do I need further steps? No. </Thought>
<Action>Done</Action>
<Final_Answer> Albert Einstein developed the theory of 
general relativity, which was published in 1915. </Final_Answer>
\end{verbatim}

\paragraph{Example 12}
\begin{verbatim}
<Question> Which Shakespeare play features the phrase 
'To be, or not to be,' and who speaks this line? </Question>
<Thought> I need to find the Shakespeare play containing 
the phrase 'To be, or not to be.' </Thought>
<Action>search_internet</Action>
<Action_Input>Shakespeare play To be, or not to be</Action_Input>
<Observation> The play is 'Hamlet.' </Observation>
<Thought> I need to find who speaks this line in 'Hamlet.' </Thought>
<Action>search_internet</Action>
<Action_Input>character who says To be, or not to be in Hamlet</Action_Input>
<Observation> The line is spoken by Hamlet. </Observation>
<Thought> Do I need further steps? No. </Thought>
<Action>Done</Action>
<Final_Answer> The phrase 'To be, or not to be' is from Shakespeare's 
'Hamlet,' and it is spoken by the character Hamlet. </Final_Answer>
\end{verbatim}

\paragraph{Example 13}
\begin{verbatim}
<Question> What is the tallest mountain in Africa, and how high is it? 
</Question>
<Thought> I need to find the tallest mountain in Africa. </Thought>
<Action>search_internet</Action>
<Action_Input>tallest mountain in Africa</Action_Input>
<Observation> Mount Kilimanjaro is the tallest mountain in Africa. </Observation>
<Thought> I need to find the height of Mount Kilimanjaro. </Thought>
<Action>search_internet</Action>
<Action_Input>height of Mount Kilimanjaro</Action_Input>
<Observation> Mount Kilimanjaro is 5,895 meters tall. </Observation>
<Thought> Do I need further steps? No. </Thought>
<Action>Done</Action>
<Final_Answer> Mount Kilimanjaro, the tallest mountain in Africa, 
is 5,895 meters high. </Final_Answer>
\end{verbatim}

\paragraph{Example 14}
\begin{verbatim}
<Question> Who was the first President of the United States to serve 
two non-consecutive terms? </Question>
<Thought> I need to find the President of the United States who served 
two non-consecutive terms. </Thought>
<Action>search_internet</Action>
<Action_Input>President who served two non-consecutive terms</Action_Input>
<Observation> Grover Cleveland was the first President to serve 
two non-consecutive terms. </Observation>
<Thought> Do I need further steps? No. </Thought>
<Action>Done</Action>
<Final_Answer> Grover Cleveland was the first President of 
the United States to serve two non-consecutive terms. </Final_Answer>
\end{verbatim}

\paragraph{Example 15}
\begin{verbatim}
<Question> What planet is the largest in our solar system, 
and what is its diameter? </Question>
<Thought> I need to find the largest planet in our solar system. </Thought>
<Action>search_internet</Action>
<Action_Input>largest planet in solar system</Action_Input>
<Observation> Jupiter is the largest planet in the solar system. </Observation>
<Thought> I need to find the diameter of Jupiter. </Thought>
<Action>search_internet</Action>
<Action_Input>diameter of Jupiter</Action_Input>
<Observation> Jupiter's diameter is approximately 139,820 kilometers. </Observation>
<Thought> Do I need further steps? No. </Thought>
<Action>Done</Action>
<Final_Answer> Jupiter is the largest planet in the solar system, 
with a diameter of approximately 139,820 kilometers. </Final_Answer>
\end{verbatim}

\paragraph{Example 16}
\begin{verbatim}
<Question> What was the first airplane to fly, and in what
year did it achieve this feat? </Question>
<Thought> I need to find the name of the first airplane to fly. </Thought>
<Action>search_internet</Action>
<Action_Input>first airplane to fly</Action_Input>
<Observation> The first airplane to fly was the Wright Flyer. </Observation>
<Thought> I need to find the year the Wright Flyer achieved 
its first flight. </Thought>
<Action>search_internet</Action>
<Action_Input>year Wright Flyer first flight</Action_Input>
<Observation> The Wright Flyer flew for the first time in 1903. </Observation>
<Thought> Do I need further steps? No. </Thought>
<Action>Done</Action>
<Final_Answer> The Wright Flyer was the first airplane to fly, 
achieving this feat in 1903. </Final_Answer>
\end{verbatim}

\paragraph{Example 17}
\begin{verbatim}
<Question> Who painted the Mona Lisa, and where is it displayed? </Question>
<Thought> I need to find the artist who painted the Mona Lisa. </Thought>
<Action>search_internet</Action>
<Action_Input>artist who painted Mona Lisa</Action_Input>
<Observation> Leonardo da Vinci painted the Mona Lisa. </Observation>
<Thought> I need to find where the Mona Lisa is displayed. </Thought>
<Action>search_internet</Action>
<Action_Input>where is the Mona Lisa displayed</Action_Input>
<Observation> The Mona Lisa is displayed in the Louvre 
Museum in Paris. </Observation>
<Thought> Do I need further steps? No. </Thought>
<Action>Done</Action>
<Final_Answer> Leonardo da Vinci painted the Mona Lisa, 
which is displayed in the Louvre Museum in Paris. </Final_Answer>
\end{verbatim}

\paragraph{Example 18}
\begin{verbatim}
<Question> Who has won the most Grand Slam tennis titles, 
and how many have they won? </Question>
<Thought> I need to find the tennis player with the most 
Grand Slam titles. </Thought>
<Action>search_internet</Action>
<Action_Input>player with most Grand Slam tennis titles</Action_Input>
<Observation> Novak Djokovic has won the most Grand Slam titles. </Observation>
<Thought> I need to find the number of Grand Slam titles he has won. </Thought>
<Action>search_internet</Action>
<Action_Input>number of Grand Slam titles Novak Djokovic</Action_Input>
<Observation> Novak Djokovic has won 24 Grand Slam titles. </Observation>
<Thought> Do I need further steps? No. </Thought>
<Action>Done</Action>
<Final_Answer> Novak Djokovic has won the most Grand Slam 
tennis titles, with 24 titles. </Final_Answer>
\end{verbatim}

\paragraph{Example 19}
\begin{verbatim}
<Question> Who was the longest-reigning monarch in British history, 
and how many years did they reign? </Question>
<Thought> I need to find the longest-reigning monarch in British history. </Thought>
<Action>search_internet</Action>
<Action_Input>longest reigning monarch in British history</Action_Input>
<Observation> Queen Elizabeth II was the longest-reigning monarch in 
British history. </Observation>
<Thought> I need to find how many years she reigned. </Thought>
<Action>search_internet</Action>
<Action_Input>length of reign Queen Elizabeth II</Action_Input>
<Observation> Queen Elizabeth II reigned for 70 years. </Observation>
<Thought> Do I need further steps? No. </Thought>
<Action>Done</Action>
<Final_Answer> Queen Elizabeth II was the longest-reigning monarch 
in British history, with a reign of 70 years. </Final_Answer>
\end{verbatim}

\paragraph{Example 20}
\begin{verbatim}
<Question> Which Shakespeare play contains the line 
\"All the world’s a stage,\" and how many years ago was it first 
performed if today is 2024? </Question>
<Thought> I need to identify the Shakespeare play with the line 
\"All the world’s a stage.\" </Thought>
<Action> search_internet </Action>
<Action_Input> Shakespeare play All the world’s a stage </Action_Input>
<Observation> The line is from \"As You Like It.\" </Observation>
<Thought> I need the year \"As You Like It\" was first performed. </Thought>
<Action> search_internet </Action>
<Action_Input> year As You Like It first performed </Action_Input>
<Observation> \"As You Like It\" was first performed in 1603. </Observation>
<Thought> I need to calculate how many years ago 1603 was from 2024. </Thought>
<Action> calculate </Action>
<Action_Input> 2024 - 1603 </Action_Input>
<Observation> 421 years. </Observation>
<Thought> Do I need to use a tool? No. </Thought>
<Action> Done </Action>
<Final_Answer> \"As You Like It\" contains the line 
\"All the world’s a stage\" and was first performed 421 years ago in 1603. </Final_Answer>
\end{verbatim}

\section{Related work}

Recent advancements in NLP have increasingly combined retrieval with generation to boost factual accuracy in knowledge-intensive tasks. Frameworks like Retrieval-Augmented Generation (RAG) dynamically retrieve and update external knowledge sources (as opposed to knowledge contained in an LLM's static, pre-stored parameters) \cite{lewis2020retrieval}. Prompting techniques such as Chain-of-Thought (CoT) \cite{wei2022chain} and ReAct \cite{yao2022react} further bolsters factuality through reasoning steps and tool access.

Closed-source, commercial search engines have arisen from the success of such retrieval techniques. Fresh off a \$500 million funding round in December, 2024, Perplexity has positioned itself as a premiere LLM-enabled search engine \cite{perplexity2024}. Other closed-source frameworks perform notably on search benchmarks, such as Exa \cite{exa} and Linkup \cite{linkup}.

While the monetary incentive for closed-source search engines is clear, opaque systems pose downsides for users and prevent researchers from building on their advances. Generative search engines are rife with bias \cite{li2024generative}, prone to hallucination \cite{venkit2024search}, amenable to surveillance \cite{zuboff2019age}, and susceptible to providing dangerous information \cite{luo2025rising}. Generative web-enabled search is a burgeoning field, catalyzed by expanding context windows and better reasoning abilities \cite{li2024survey}; with such growth comes a responsibility to democratize access and mitigate risks with open-source research.

Some open-source efforts have made commendable strides in the desiging LLM-assisted search architectures. \cite{kovsprdic2024verif} create an information retrieval system to access pre-indexed scientific articles with the assistance of a fine-tuned LLM. \cite{neague2025semantica} construct a new, fully decentralized database using LLM-derived embeddings (rather than accessing an existing web index) for scalable semantic search by routing queries through semantically similar neighbors.

Recent work has also focused on actively querying the web, such as \cite{chen2024mindsearch}, which harnesses a multi-agent LLM framework to incrementally plan, coordinate, and integrate large volumes of live web information in real time. Other work has brought a multimodal dimension to web-enabled search for real-time interaction with live websites \cite{zheng2024gpt}.

\end{document}